\documentclass[pdflatex,sn-mathphys-num]{sn-jnl}

\usepackage{graphicx}%
\usepackage{multirow}%
\usepackage{amsmath,amssymb,amsfonts}%
\usepackage{amsthm}%
\usepackage{mathrsfs}%
\usepackage[title]{appendix}%
\usepackage{xcolor}%
\usepackage{textcomp}%
\usepackage{manyfoot}%
\usepackage{booktabs}%
\usepackage{algorithm}%
\usepackage{algorithmicx}%
\usepackage{algpseudocode}%
\usepackage{listings}%

\usepackage{adjustbox}

\theoremstyle{thmstyleone}%
\theoremstyle{thmstyletwo}%

\theoremstyle{thmstylethree}%

\begin{document}

\title[Instance Discrimination for Link Prediction]{Instance Discrimination for Link Prediction}


\author[1]{\fnm{Valentin} \sur{Cuzin-Rambaud}}\email{valentin.cuzin-rambaud@univ-lyon1.fr}

\author[1,2]{\fnm{Mathieu} \sur{Lefort}}\email{mathieu.lefort@univ-lyon1.fr}

\author[1]{\fnm{Rémy} \sur{Cazabet}}\email{remy.cazabet@univ-lyon1.fr}

\affil[1]{Université Lyon 1, INSA Lyon, CNRS, LIRIS, UMR 5205, Lyon, France}
\affil[2]{Université Rennes, Inria, CNRS, IRISA - UMR 6074; F-35000 Rennes, France}

\abstract{
Recently, instance discrimination models have emerged as a major solution for self-supervised learning. Having already demonstrated its effectiveness in the image domain, instance discrimination learning is now proving equally convincing in the graph domain, in particular for node classification. However, fewer contributions have tackled the link prediction task. In this contribution, we propose to adapt existing methods to this context. We first provide a rigorous evaluation of existing self-supervised models in the field of link prediction, showing that the main performance depends on the augmentation process (like in computer vision). We then propose a new structural augmentation based on the community structure that is relevant for link prediction. Our main contribution introduces two new models, L-GRACE and L-BGRL, based on link representations instead of node representations, which improve the performance of the existing methods, especially on unattributed graphs, and we show that they perform on par with the state of the art, both in supervised and self-supervised contexts.
}


\maketitle

\section{Introduction}
The link prediction task is an important field in which the goal is to identify missing links in networks, such as infrastructure, biological, social, and collaboration networks \cite{kumar2020link}. Beyond multiple applications such as recommendations (e.g., users/products) and healthcare (e.g., discovering new interactions between proteins), studying link prediction is a way to understand the structure of networks, i.e., the underlying principles of their organization. The task can be applied to one unique graph (transductive setting) or to a collection of graphs, some used in training and others for the prediction (inductive setting).

\textbf{Problem statement.} An undirected graph $G$ is defined by the quadruplet $(V,E,A,X)$, where $V$ is a set of $n$ nodes, $E$ is a set of edges, $A$ is the adjacency matrix, with $A_{uv} = 1, \iff (u,v) \in E$ and reciprocally $A_{uv} = 0, \iff (u,v) \notin E$, $X$ is a node feature matrix, $X_u$ being the feature vector of node $u$. Our work focuses on unattributed graphs, when $X$ is taken as the identity matrix $I$ of size $n*n$. We denote a positive link as a pair of nodes $(u,v) \in E$ simply noted $uv$ and a negative link as a pair of nodes $(u, v) \notin E$ noted $\bar{uv}$ in this paper. Given an unlabeled/masked pair of nodes $(u,v)$, the task of transductive link prediction is to determine if the link is positive $(u,v) \in E$ or negative $(u,v) \notin E$.

From the structure of the graph and potentially useful node attributes, Graph Neural Networks (GNN) \cite{scarselli2008graph} based models appear to be the most effective method for link prediction \cite{kipf2016semi,kipf2016variational}. Furthermore, Self-Supervised Learning (SSL), and especially instance discrimination methods, show excellent performance on node or graph classification tasks and are competitive with supervised models. Instance discrimination learning comes from the image research field, with well-known methods such as SimCLR \cite{chen2020simple}, BYOL \cite{grill2020bootstrap}, and so forth. The principle is to generate two alternative instances, also known as views, through augmentations of the same input data, and then learn to maximize the similarity between the representations of these two instances, relying on the mutual information between them. The model constrains two similar synthetic data instances to have similar representations. A well-known pitfall of these methods is the model collapse, where all data obtain representations that are too similar. We focus in this work on two families of solutions to this problem: Contrastive Learning, where we learn the dissimilarity between pairs that are not similar, and Asymmetric Learning, with the use of asymmetric encoders. Previous works have used these approaches in the context of graph classification and node classification tasks, i.e., using graph contrastive learning (GCL) \cite{you2020graph,zhu2020deep,zhu2021graph}, as well as asymmetric methods \cite{thakoor2021large,bielak2022graph}. However, few works have been published on the link prediction task.
The success of graph contrastive learning models in node classification hinges largely on the effectiveness of the augmentation process \cite{zhu2021graph}. While instance discrimination models for node classification have proven their effectiveness, they were not designed with the link prediction task in mind. In this article, we explore how to adapt them to the problem of link prediction. Hence, in this work, we answer the main question: 

\textbf{How to adapt instance discrimination models to the link prediction task in graphs?}

To answer this question, we evaluate some instance discrimination methods for link prediction in a transductive setting against recent supervised link prediction baselines. 
We first demonstrate that the model's performance is strongly reliant on the augmentation strategy and then propose a new augmentation method using an SBM generator to generate new instances based on the community structure, thereby learning more effective representations. Secondly, we propose a custom loss designed for link prediction. We use it to propose alternatives to the well-known GRACE and BGRL methods, named L-GRACE and L-BGRL.

The paper is organized as follows: In section \ref{related work}, we introduce the most relevant state of the art in link prediction and instance discrimination methods. We then present our contribution in two steps, first focusing on proposing new augmentations tailored for link prediction \ref{adapted augmentation}, and then a custom loss function \ref{adapted model}. Finally, we demonstrate the interest of our propositions in section \ref{experiments}.

\section{Related Work} \label{related work}

We will present related works regarding two axes: link prediction models and self-supervised learning approaches.

\subsection{Methods for Link prediction}

Several historical methods exist for link prediction based on heuristics (unsupervised). Some of them are based on local structural features like Common Neighbors \cite{newman2001clustering}, Adamic Adar \cite{adamic2003friends}, and Resource Allocation \cite{zhou2009predicting}, while others are based on global heuristics such as Katz \cite{katz1953new}, or SimRank \cite{jeh2002simrank}. Even if these statistical methods are old, their predicted link sets do not fully overlap with those predicted by GNN models \cite{mao2023revisiting}. Thus, recent methods try to approximate these statistical methods by integrating structural features to improve the expressiveness and performance of the model \cite{zhang2018beyond,yun2021neo,chamberlain2022graph,wang2023neural,dong2024pure}.

One of the first frameworks that combines GNNs and structural features for link prediction is \textit{Subgraphs Embeddings and Attributes for Link Prediction} (SEAL) \cite{zhang2018link}. The method extracts the local subgraph surrounding a target node pair and treats that pair as the subgraph's center. Each node in the extracted subgraph is then labeled according to its distance to the center. These distance-based node labels permit learning about the local structural features. The main challenge of this technique is the computational cost of labeling, which is necessary at inference. To solve this problem and keep the good performances of SEAL, \textit{Efficient Link Prediction with Hashing} (ELPH) and BUDDY \cite{chamberlain2022graph}, or \textit{Message Passing Link Predictor} (MPLP) \cite{dong2024pure} proposed improved frameworks leveraging the subgraph features and improving link prediction accuracy. Finally, \textit{Neural Common Neighbor with Completion} (NCNC) \cite{wang2023neural} utilizes the common neighbors information by combining the embeddings of common neighbors of the node pair to evaluate with the embeddings of the node pair itself.
All of these methods are supervised, in the sense that they use the presence or the absence of a link as a label for a node pair, and they define an objective loss, such as the binary cross-entropy (BCE), to optimize.

\subsection{Self-supervised learning methods}
Several methods have been proposed for self-supervised learning (SSL) on graphs, although none of them is dedicated to the link prediction task. 
The objective of these methods is to yield embeddings that can later be used for various tasks, most of them evaluated on node or graph classification.

GRACE \cite{zhu2020deep} is a graph contrastive learning (GCL) method, mainly used for node classification. The principle of graph contrastive learning is to generate variations of an observed graph, and then to learn a node representation that is similar for the original graph and its variation. The key part of such an approach is the process by which variations of the original graph are generated. In GRACE, the augmentation process consists of introducing random perturbations of the graph, namely removing edges and masking node attributes. The loss is computed through Noise-Contrastive Estimation based function (infoNCE) \cite{oord2018representation}, which combines intra-view and inter-view similarities. Hence, for each positive pair (the same node in both views) ($u_i, v_i$), the loss is defined in equation \ref{eq:grace_loss}

\begin{equation}
    \ell(u_i, v_i) = \log \frac{e^{\theta(u_i, v_i)/\tau}}{ \underbrace{e^{\theta(u_i, v_i)/\tau}}_{\text{the positive pair}} + \underbrace{\sum_{k=1}^{N} \mathbf{1}_{[k \neq i]} e^{\theta(u_i, v_k)/\tau}}_{\text{inter-view negative pairs}} + \underbrace{\sum_{k=1}^{N} \mathbf{1}_{[k \neq i]} e^{\theta(u_i, u_k)/\tau}}_{\text{intra-view negative pairs}}}
    \label{eq:grace_loss}
\end{equation}

with $\theta(u_i, v_i) = s(g(u_i), g(v_i))$, where $s$ is the cosine similarity function, and $g$ is a non-linear projection head (namely, an MLP with 2 layers). The goal is to maximize inter-view similarity between two nodes of the positive pairs, while maximizing distance to other nodes in the intra-view and the same node in inter-view. Finally, the objective function to maximize is equation \ref{eq:grace_obj}.

\begin{equation}
     \mathcal{J} = \frac{1}{2N} \sum_{i=1}^{N} \left[ \ell(\mathbf{u}_i, \mathbf{v}_i) + \ell(\mathbf{v}_i, \mathbf{u}_i) \right]
    \label{eq:grace_obj}
\end{equation}

Another work in the continuity of GRACE is \textit{Graph Contrastive learning with Adaptive augmentation} (GCA) \cite{zhu2021graph}, which introduces an augmentation adaptive to the importance of nodes in the graph\label{aug}. Using three different centrality measures, including degree centrality, eigenvector centrality, and PageRank centrality of nodes in the graph, the GCA model reaches better performance than the original GRACE on node classification. After computing the centrality of each node, the augmentation process favors removing edges that link less important nodes and masking more attributes of less important nodes. The improvements brought by GCA highlight the importance and potential of the augmentation process.

Several contributions proposed to use community structure to improve GCL. \textit{gCooL} \cite{li2022graph} is the first work proposing to combine community information with GCL, with a focus on the augmentation process and the model itself. \textit{Community-Strength-Enhanced Graph Contrastive Learning} (CSGCL) \cite{chen2023csgcl} is a more recent work bringing the density of community information to the GCL model, and showing good results on node and graph classification. As with GCA, CSGCL proposes new augmentation adaptive to the community structure: \textit{Communal Attribute Voting} (CAV) and \textit{Communal Edge Dropping} (CED). CAV tends to preserve attributes that are more influential in strong communities, while CED tends to preserve edges intra-community more than inter-community and favors conserving edges from strong communities more than weak ones. 
Moreover, CSGCL brings an important upgrade to the InfoNCE loss: in the similarity computation between 2 nodes, it adds the respective community strengths of both nodes, refining the node representation in the context of its community. Similarly, Community-Invariant Graph Contrastive Learning (CI-GCL) \cite{tan2024community} learns to create community-invariant augmentations through learnable augmentation that maximizes spectral changes in the Laplacian matrix. The method is evaluated for the graph classification task, which differs more from node classification or link prediction.

\textit{Bootstrapped Graph Latents} (BGRL) \cite{thakoor2021large} is an adaptation of BYOL \cite{grill2020bootstrap} from images to the graph case. It is an asymmetric method that is really fast and scalable on large graphs, because it does not need to select negative nodes like contrastive methods do, which is memory-intensive. The principle is the use of two neural network encoders that learn at different speeds. One of them, called the “online encoder”, updates its weights with the cosine similarity between the final representations of two instances, while the other one, called the “target encoder”, learns with an Exponential Moving Average (EMA) from the weights of the online encoder. Thus, the target encoder weights are based on the online encoder, creating a learning gap between the two encoders.
For the augmentation process, BGRL performs the same random perturbations as GRACE. The authors observe that adaptive augmentation, such as GCA, does not bring an upgrade in performance with their architecture. In our evaluation, we will re-examine this result for link prediction and demonstrate that BGRL benefits from well-chosen adaptive augmentation.

A recent study focusing on link prediction with instance discrimination learning \cite{shiao2022link} reveals that BGRL performs better than several other methods on link prediction. They are presenting a new model named Triplet-BGRL (T-BGRL) based on BGRL that modifies the loss by maximizing the distance between the augmentation and corruption representations (with corrupted data). This corrupted representation is viewed as a negative augmentation, complementing the asymmetry mechanism to reduce the risk of overfitting in an inductive setting. The obtained results show that T-BGRL is not relevant in a transductive setting. Compared with this study, which only focuses on changing the loss function, our work focuses on the augmentation process and the loss function.

Several other methods exist in instance discrimination for graphs \cite{liu2022graph}, like Graph-Barlow-Twins that adapt Barlow-Twins \cite{zbontar2021barlow} to graphs, or graphCL \cite{you2020graph}, but they don't bring major improvement, or special treatment for instance discrimination with graphs. So we have based our new models on GRACE and BGRL as they remain the standard baseline in the domain, while belonging to the two main different families of methods to avoid model collapse, the first using Contrastive Learning, the second Asymmetric Learning.

Finally, some works use the line graph transformation to transform the problem of link prediction to a binary classification node, as does LGLP \cite{cai2021line} for a supervised link predictor or even applying a contrastive framework on a line graph, as does LGCLP \cite{zhang2023line}. The line graph transformation has high time complexity and does not scale to large graphs. The complexity is $\mathcal{O}(n^2)$ with $n$ edges in the graph. This issue led our research to avoid using the line graph transformation.

\section{Proposed Approaches: Adapted instance discrimination model for link prediction} \label{contrib}

The objective of an instance discrimination model is to learn representations that are invariant to the augmentations. Models already introduced in the literature were designed mostly with node classification in mind, and thus create node representations using a loss function comparing node embeddings. However, in a link prediction setting, the basic unit used for prediction is pairs of nodes, seen as combined embeddings, considering only individual node embeddings is a potential loss of information.

To adapt the instance discrimination model for link prediction, we consider the two main parts contributing to the final performance of the model: in section \ref{adapted augmentation}, we present our contribution to the augmentation process, and in section \ref{adapted model}, we present our contribution to the loss function. 

Our contribution focuses on non-attributed graphs for the transductive link prediction task, where the model discovers and exploits knowledge solely on the structural information in one graph to predict links.

\subsection{SBM-based augmentation} \label{adapted augmentation}

The augmentation process is critical in an instance discrimination model: selecting the transformations to which the model should be invariant directs learning toward the specific graph structure of interest. 
Our first contribution focuses on a new augmentation capable of bringing community information to the learning process. 

Community structure is known to be useful in link prediction settings \cite{ghasemian2020stacking}. A community represents a group of nodes that is more connected than the rest of the graph. Thus, knowing communities helps to identify groups of nodes between which links are more likely to appear. The Stochastic Block Model (SBM) \cite{karrer2011stochastic} is a statistical graph generative model where nodes are grouped in blocks, each pair of blocks having its own probability of creating edges between its nodes. SBM can be fitted to real graphs, typically using Bayesian inference \cite{peixoto2014efficient}. An SBM generator fitted with a given partition of a graph generates random graphs with the same number of nodes as the reference, similar inter-block densities, but an alternative set of links. SBMs fitted to a graph are known to be efficient tools for link prediction \cite{ghasemian2020stacking}.

We propose to use SBMs to generate graph augmentations. The idea is that individual links between people (A being friends with B and not C) are noise, unreliable information, while the community structure, being the pattern driving the creation of these links, is more reliable. For instance, imagine students splitted in classes in a high school. The details of which student becomes friends with which student is something noisy, and if we could replay the same configuration twice, different friendships within the same class might appear, due to sensibility on minor details in the first days, e.g., who sits close to whom in class on the first days, etc. However, we can predict with a strong probability that most of the friendship links that will appear will do so \textit{inside} the class, thus driven by the community structure.

\begin{figure}[h!t]
    \centering
    \includegraphics[width=0.75\linewidth]{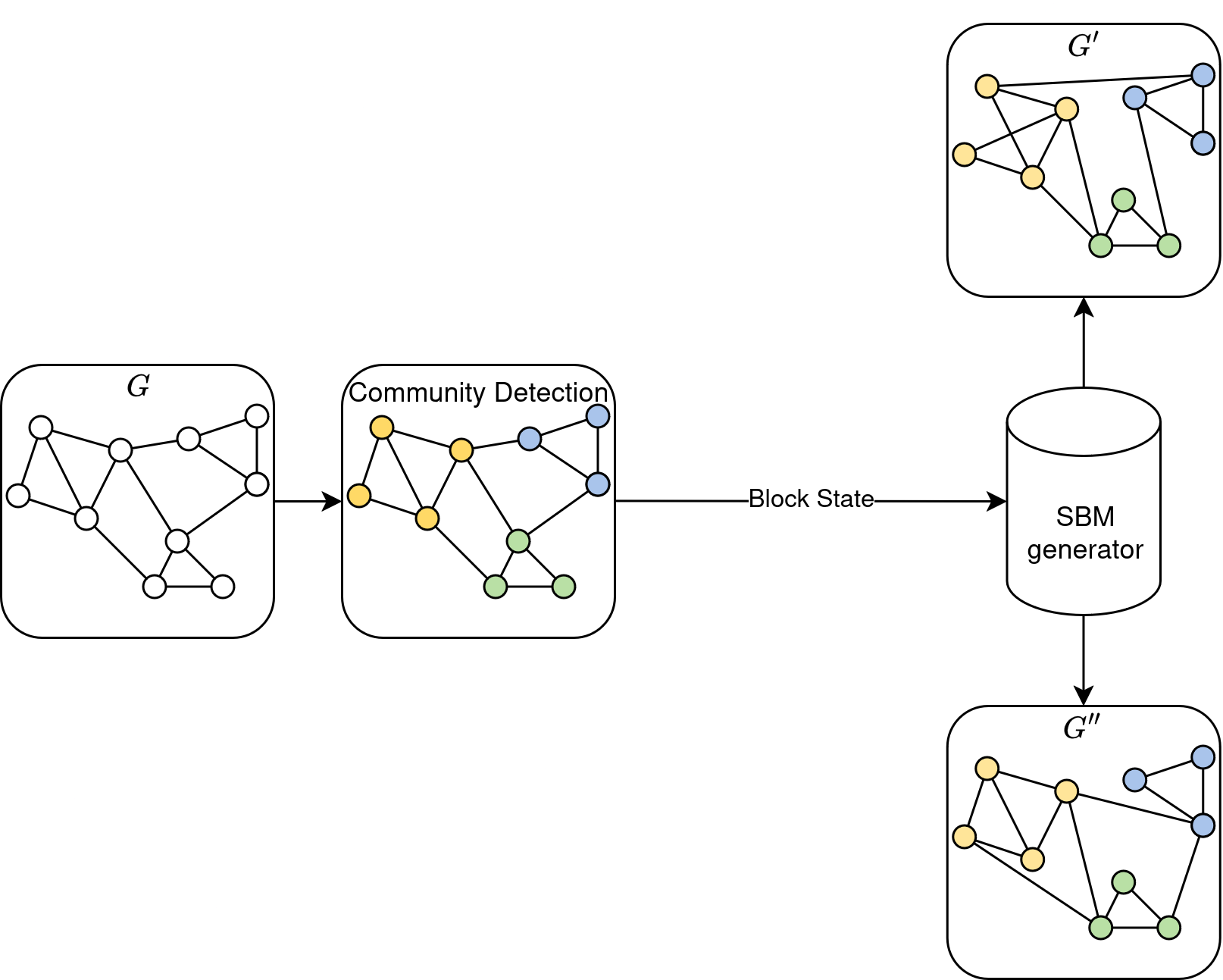}
    \caption{The SBM augmentation process. Using a simple algorithm of community detection, like Louvain, Leiden, and Infomap, for obtaining the Block State of the graph. Then the SBM generator creates a new graph $G'$ that respects the block state of the input graph $G$.}
    \label{fig:sbm}
\end{figure}

Our proposed pipeline is executed in two main steps: 1) inferring the block model from the input graph, then 2) generating graphs during the training (see Fig.\ref{fig:sbm}). It can be applied to whichever model.
To infer communities, we have selected three classical community detection algorithms: Louvain, Leiden, and Infomap, and fixed the algorithm as a hyperparameter of the augmentation. Note that we did not use SBM inference methods for performance reasons, those methods remain much less efficient. This limit could be overcome in future works. Secondly, we give the block state (which assigns a block to each node) to the SBM generator. Finally, the SBM generator can generate many graphs respecting the same block state, but with different edges between nodes. 
We design two alternatives of the model: when providing two graphs to the model, either 1) one instance is the original graph while the other is the result of the augmentation process, or 2) both instances are the result of an augmentation process.

\subsection{Contrastive Loss Function for Link Prediction} \label{adapted model}

Although applicable to link prediction, instance discrimination models applied to graphs were designed for node or graph classification tasks. In GRACE or BGRL, the loss consists of making node representations similar, which is relevant for node-centric tasks. We propose instead a loss function tailored for link prediction.

The main idea of this loss is to use link representations instead of node representations as atomic units.
Using an MLP, we defined the representation of a link $H_{uv}$ as the projection of the Hadamard product of the embedding $H_u$ and $H_v$ of a node pair.

\begin{equation}
    H_{uv} = MLP(h_u \odot h_v)
    \label{eq:link_rpz}
\end{equation}

\begin{figure}[h!t]
    \centering
    \includegraphics[width=0.9\linewidth]{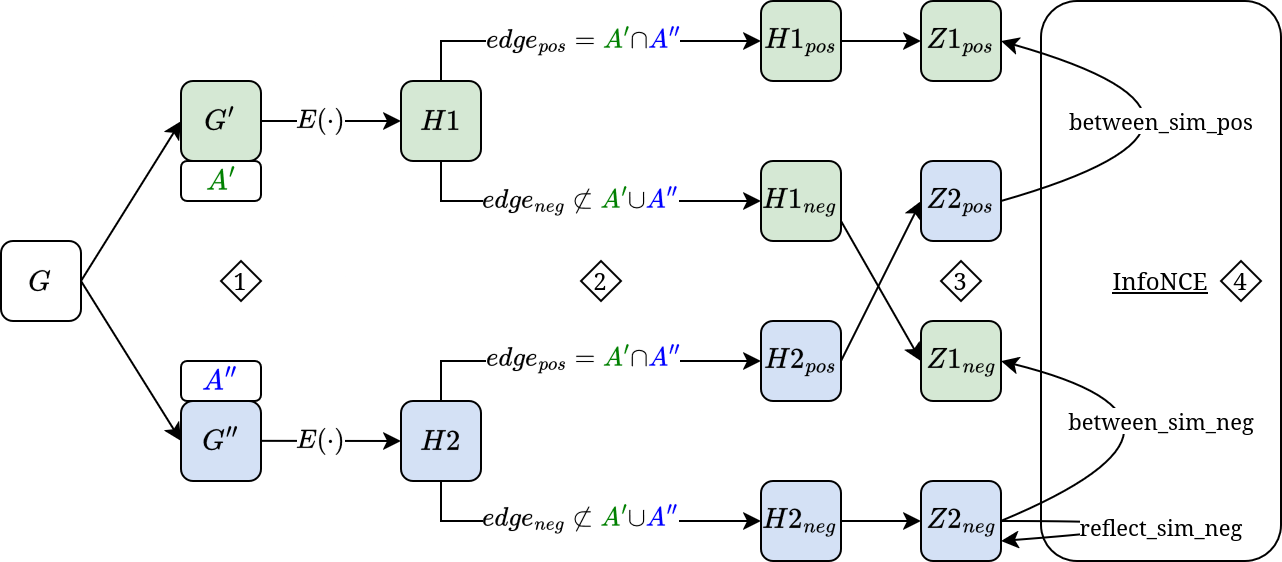}
    \caption{L-GRACE: Our link-based contrastive model for link prediction. The embedding (obtained by $E(\cdot)$) of nodes is combined through negative and positive links, then a projection head creates the final embedding of links $Z$. We apply the InfoNCE loss, trying to maximize similarities between positive links embedding in the 2 views $Z1_{pos}$ and $Z2_{pos}$ and maximize dissimilarities for negative links embedding in the self-view and between views.}
    \label{fig:L-GRACE}
\end{figure}

We present first Link-GRACE (L-GRACE) as the contrastive model using the link representation (see Fig. \ref{fig:L-GRACE}). Like GRACE, we begin the process with the two augmented graphs $G'$ and $G''$ and their adjacency matrices $A'$ and $A''$ respectively (see Fig \ref{fig:L-GRACE} step 1). After obtaining embeddings $H1$ and $H2$ using a GCN encoder, we create link representations for both the positive node pairs present in the two instances ($edge_{pos} = A'\cap A''$) and a sample of negative node pairs, not present in the two instances ($edge_{neg} \not\subset A'\cup A''$), with the same size as $edge_{pos}$ (see Fig \ref{fig:L-GRACE} step 2). This particular point is responsible for learning the representation of links that are positive by maximizing the mutual information between views, while differentiating from all other negative links. To obtain link representations, we apply equation \ref{eq:link_rpz} systematically to each node pair retained in sets $edge_{pos}$ and $edge_{neg}$ from $H1$ and $H2$, which permits us to access four 
final sets of link representations: $Z1_{pos}$, $Z1_{neg}$, $Z2_{pos}$, and $Z2_{neg}$ (see Fig \ref{fig:L-GRACE} step 3). Finally, we update the model's weights using a modified InfoNCE-based objective function (see Fig \ref{fig:L-GRACE}, step 4, and the associated eq. \ref{eq:L-GRACE-loss}).

When computing the loss, we maximize the similarity between the positive node pair representations in the two views 
(between\_sim\_pos), while maximizing the dissimilarities with other negative node pair representations intra-view (reflect\_sim\_neg) and inter-view (between\_sim\_neg). Formally, we define:
\begin{itemize}
    \item $uv^j_i \in Zj_{pos}$ the final representations of the $i$-th positive link in the $j$ view
    \item $\bar{uv}^j_k \in Zj_{neg}$ the final representations of the $k$-th negative link in the $j$ view
    \item $\theta(\cdot, \cdot)$ the cosines similarity function
    \item $\tau$ denotes a temperature parameter
    \item $N$ the total number of negative links
\end{itemize}
for the use of equation \ref{eq:L-GRACE-loss}, which computes the loss for a link representation in both views.

\begin{equation}
\begin{split}
    \ell(uv^1_i, uv^2_i) = \log \frac{e^{{\theta(uv^1_i, uv^2_i)/\tau}}}{ \underbrace{\sum_{k=1}^{N} e^{\theta(\bar{uv}^1_i, \bar{uv}^2_k)/\tau}}_{\text{between\_sim\_neg}} + \underbrace{\sum_{k=1}^{N} \mathbf{1}_{[k \neq i]} e^{\theta(\bar{uv}^1_i, \bar{uv}^1_k)/\tau}}_{\text{reflect\_sim\_neg}}}
\end{split}
\label{eq:L-GRACE-loss}
\end{equation}

At each epoch, the model learns to maximize inter-view similarity between a positive link representation in both views, and distinguishes it from all other negative link representations in $edge_{neg}$ from intra-view and inter-view.
Finally, the objective function to maximize is defined as the average over all positive link pairs, as specified in equation \ref{eq:L-GRACE-objectif}. We apply the loss in both directions.

\begin{equation}
    \mathcal{J} = \frac{1}{2N} \sum_{i=1}^{N} [ \ell(uv^1_i, uv^2_i) + \ell(uv^2_i, uv^1_i) ]
    \label{eq:L-GRACE-objectif}
\end{equation}

\begin{figure}[h!t]
    \centering
    \includegraphics[width=0.9\linewidth]{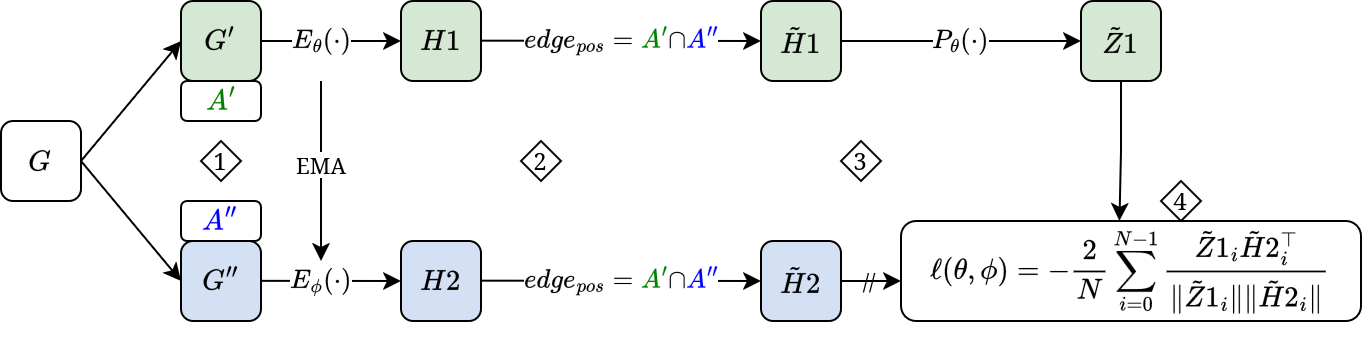}
    \caption{L-BGRL: Our link-based asymmetric model for link prediction. With two encoders that do not share the same weights, we learn as the BGRL model does. We create the embedding of a node pair only when nodes are in relation, with a positive link.}
    \label{fig:L-BGRL}
\end{figure}

Secondly, we present L-BGRL, our adaptation of BGRL for an asymmetric model of link prediction. As with BGRL, we create two augmented views $G'$ and $G''$ which are encoded respectively by $E_\theta$, the online encoder, and $E_\phi$, the target encoder (see Fig.\ref{fig:L-BGRL} step 1). Then, we obtain embeddings $\tilde{H}1$ and $\tilde{H}2$ with positive links in both instances $A' \cap A''$, combined by a Hadamard product of node embeddings (Eq.\ref{eq:link_rpz}) in $H1$ and $H2$ (see Fig.\ref{fig:L-BGRL} step 2). The embedding $\tilde{H}1$ (with the online encoder) is transformed by an MLP projection $P_\theta$ to obtain the final representation $\tilde{Z}1$ and the embedding $\tilde{H}2$ is considered as the final representation, which is used in the loss function (see Fig.\ref{fig:L-BGRL} step 3). 

\begin{equation}
    \ell(\theta, \phi) = -\frac{2}{N} \sum_{i=0}^{N-1} \frac{\tilde{Z}1_{i} \tilde{H}2_{i}^\top}{\|\tilde{Z}1_{i}\| \|\tilde{H}2_{i}\|}
    \label{eq:L-BGRL-loss}
\end{equation}

Finally, we apply the loss function (see Fig.\ref{fig:L-BGRL} step 4, Eq.\ref{eq:L-BGRL-loss}) in both directions, to update the weights $\theta$ of the online encoder, while the weights $\phi$ are updated with an Exponential Moving Average (EMA), as in BGRL.

\section{Experiments} \label{experiments}

In this section, we are going to compare five Instance discrimination models (GRACE, CSGCL, BGRL, L-GRACE, L-BGRL) against several supervised models and scores from the T-BGRL study. We also test the impact of introducing our augmentations in the different models independently.

Our setup is explained in the following, and the code is publicly available at \href{https://github.com/valentincuzin/GCL-Link-Prediction}{https://github.com/valentincuzin/GCL-Link-Prediction}.

\subsection{Experimental Setup}

\subsubsection{Dataset}

The datasets used are a combination of those from the MPLP \cite{dong2024pure} and T-BGRL studies \cite{shiao2022link}, and permit the evaluation of instance discrimination methods in various domains. Our evaluation focuses on unattributed graphs, as our contribution intends to discover knowledge from the structure of the graph. We have eight non-attributed graphs: the USAir \cite{batagelj2006pajek}, Power \cite{watts1998collective} and Router \cite{spring2002measuring} infrastructures networks, the Yeast \cite{von2002comparative}, Celegans \cite{watts1998collective} and Ecoli \cite{zhang2018beyond} biological networks, the NS collaboration network \cite{newman2006finding} and the PB web pages graph \cite{ackland2005mapping}. Moreover, we extend our analysis to six attributed graphs: the Cora and Citeseer \cite{yang2016revisiting} citation networks, the Coauthor-CS and Coauthor-Physics coauthorship networks, and the Amazon-Computers and Amazon-Photo co-purchase networks \cite{shchur2018pitfalls}. Every graph is static, undirected, and evaluated in a Transductive setting, which means evaluating link prediction in one graph by masking positive edges. The table \ref{tab:benchmark} resumes the datasets' statistics. For non-attributed graphs, we have set a unique ID for every node following the identity matrix, so $X=I$.
For the proportion of the random link split, we test two different configurations: 70\% train, 10\% validation, and 20\% test for non-attributed graphs (as in MPLP), and 85\% of train, 5\% of validation, and 10\% of test for attributed graphs (as in T-BGRL).

\begin{table*}[h!t]
    \centering
    \caption{Statistics of benchmark datasets}
    \resizebox{\textwidth}{!}{
    \begin{tabular}{llllll}
        \toprule
        Datasets & \#Nodes & \#Edges & Average and std node degree & max degree & Attribute Dimension \\
        \midrule
        cora & 2 708 & 10 556 & 3.89$\pm$5.22 & 168 & 1 433 \\
        citeseer & 3 327 & 9 104 & 2.73$\pm$3.38 & 99 & 3 703 \\
        cs & 18 333 & 163 788 & 8.93$\pm$9.11 & 136 & 6 805 \\
        physics & 34 493 & 495 924 & 14.38$\pm$15.57 & 382 & 8 415 \\
        computers & 13 752 & 491 722 & 35.76$\pm$70.31 & 2992 & 767 \\
        photo & 7 650 & 238 162 & 31.13$\pm$47.28 & 1434 & 745 \\
        USAir & 332 & 4 252 & 12.81$\pm$20.13 & 139 & - \\
        NS & 1 589 & 5 484 & 3.45$\pm$3.47 & 34 & - \\
        PB & 1 222 & 33 428 & 27.36$\pm$38.42  & 351 & - \\
        Yeast & 2 375 & 23 386 & 9.85$\pm$15.50 & 118 & - \\
        Celegans & 297 & 4 296 & 14.46$\pm$12.97 & 134 & - \\
        Power & 4 941 & 13 188 & 2.67$\pm$1.79 & 19 & - \\
        Router & 5 022 & 12 516 & 2.49$\pm$5.29 & 106 & - \\
        Ecoli & 1 805 & 29 320 & 16.24$\pm$48.38 & 1030 & - \\
        \bottomrule
    \end{tabular}
    }
    \label{tab:benchmark}
\end{table*}

\subsubsection{Metrics}

To train the link predictor head, we iteratively sample a batch of edges in the train split that are masked, and then, we sample an equal number of negative edges. The model tried to predict on these masked pairs of nodes, where positive links exist. 
Several metrics are commonly used for evaluating link prediction.
Here, we followed the choice of T-BGRL and MPLP to use Hits@50. We also present ROC-AUC and AP scores in the appendices. Hits@k in this context is defined as the mean of the number of good predictions $y_{pos}$ which have a score greater than the $k$ worst prediction $y_{neg}$: 

\begin{equation}
    \text{hits@k} = \frac{1}{|y_{pos}|} \sum_{i}{(y_{pos}[i]>sort(y_{neg})[-k])}
\end{equation}

Every score is obtained using 10 runs to train the model on 10 different random splits of the same graph. We report mean and standard deviation as means$\pm$std in tables. On each run, we set a different seed for reproducibility, which takes effect in every random sub-process, like community-detection, initial link splitting in train, validation, and test split.
For a set of results on a graph (Tab. \ref{tab:non-attr} and in the appendix), we identify groups of significantly best and significantly worst methods using a Friedman test \cite{friedman1937use} to rank methods and obtain a p-value, followed by a Bonferroni-Dunn test \cite{dunn1961multiple} to obtain best and worst groups. Significant best methods are annotated with a star $\bigstar$ and significant worst are annotated with a cross X.

\subsubsection{Models and Hyperparameters}

We based our experimental setup on the T-BGRL setup for a fair comparison. For that reason, we select the BGRL encoder and a Hadamard product MLP decoder for our models. We first train the encoder and freeze its weights before training the decoder. As a supervised baseline, we also train the encoder and decoder jointly, as reported in "GCN" in the table.
The encoder is composed of one to four GCN convolutional layers, followed by a normalization layer and a PReLU activation function. 
The MLP decoder consists of two linear layers with a ReLU activation between them. We apply a sigmoid function to the output of the last layer to obtain the final prediction score.

For the SBM generator, we use the implementation of graph tools, as it generates graphs faster than Networkx's implementation. We initialize the BlockState with the previously discovered graph partition with any community detector, then we use the micro-canonical model to sample the graph, without allowing self-loops or parallel edges generation.

We fixed the number of epochs for the MLP decoder to 100, while the encoder's number of epochs is tuned as a hyperparameter.
We made 25 runs of hyperparameter tuning using the Optuna \cite{optuna_2019} framework and its implementation of the Tree-structured Parzen Estimator (TPE), for all models and augmentations per dataset. The hyper-parameter search made by Bayesian Optimization is better than traditional grid-search. We present all parameters that we have tuned in table \ref{tab:hp}.

\begin{table*}[h!t]
    \centering
    \caption{explanation of parameters that we have tuned for each model.}
    \resizebox{\textwidth}{!}{
    \begin{tabular}{llll}
    \toprule
        Name & type & search space & description \\
    \midrule 
        ct epochs & int & \{100, 500, 1500, 3000\} & number of epochs for training \\
        batch size & int & [256, 6400], 64 step & size of the batch \\
        GNN lr & float & [0.0001, 0.01] & learning rate of the GNN for the model encoder \\
        proj hidden & int & [64, 512], 64 step & hidden layer's size in MLP projection of the model \\
        loss func & str & {'log\_sig', 'bce'} & choice between bce or log sigmoid loss function \\
        pred lr & float & [0.0001, 0.01] & learning rate of the model decoder \\
        mask input & bool & {True, False} & masking or not the input at the training \\
        weight decay & float & [1e-6, 1e-4] & the weight decay during training optimization \\
    \midrule
        n layers & int & [1, 4] & numbers of layers for the encoder \\
        layer size & int & [64, 512], 64 step & number of neurons per layers \\
        batch layer norm & bool & {True, False} & batchNorm or LayerNorm in the encoder \\
        batchnorm mm & float & [0.8, 1.0] 0.01 step & momentum for the batchNorm case \\
        weight standardization & bool & {True, False} & standardize weight at each epoch \\
        tau & float & [0.1, 0.9] 0.1 step & tau for GRACE based model training \\
    \midrule
        drop edge rate 1 & float & [0.0, 0.9] 0.1 step & drop edge rate for instance 1 \\
        drop edge rate 2 & float & [0.0, 0.9] 0.1 step & drop edge rate for instance 2 \\
        drop feature rate 1 & float & [0.0, 0.9] 0.1 step & drop feature rate for instance 1 \\
        drop feature rate 2 & float & [0.0, 0.9] 0.1 step & drop feature rate for instance 2 \\
        commu detect & str & {'louvain', 'leiden', 'infomap'} & community detection algo for sbm augmentation \\
        
    \bottomrule
    \end{tabular}
    }
    \label{tab:hp}
\end{table*}

Every run has been performed on at least an NVIDIA V100 32GB GPU or an NVIDIA H100 96GB for the largest graph datasets.

\subsubsection{Notation of Augmentations}

Since GRACE and GCA are strictly equivalent models, differing only through their augmentation strategy, we reported them as GRACE in the results table and specified the augmentation strategy used. To designate the various augmentations, we will use the following abbreviations: (1) "deg" is the Degree centrality adaptive augmentation, (2) "evc" is the EigenVector centrality adaptive augmentation, and (3) "pr" is the PageRank centrality adaptive augmentation. The classical augmentation process of GRACE is thus named "random", without adaptive selection of edges/node attributes to preserve.
In the same way, we clearly separate the architecture used in CSGCL from its augmentation, which includes attributes and edges tricks to preserve community structure, shortening "scom" for the strength community augmentation(e.g. "BGRL scom"). We systematically evaluate all the augmentation configurations on each model. 
Our augmentation process using the SBM generator \ref{adapted augmentation} is tested in two ways: "sbm" is the augmentation on only one instance, with the unchanged input graph on the second instance, and "sbm2" is the augmentation on both instances \ref{adapted augmentation}. We name "optim" augmentation, the best adaptive augmentation for a model among \{deg, evc, pr, scom, sbm, and sbm2\}.

\subsection{Results}

In this section, we first analyze the effect of the augmentation process, and then of L-GRACE and L-BGRL, our adaptations of GRACE and BGRL using link representation.

\subsubsection{Influence of the Augmentation}

\textbf{SBM Augmentation proof of concept.} To verify the idea behind the SBM Augmentation, we first test an "oracle" version in which we perform the community detection before doing the random split in train, validation, and test, so that the block state doesn't depend on the random splitting. The Table \ref{tab:sbm_oracle} shows the result of this experiment. 
\begin{table*}[h!t]
    \centering
    \caption{Link prediction results with Hits@50 metric, of the oracle version of sbm augmentation, on non-attributed graphs. Supervised methods are reported (sup). The top three scores per column are colored by {\color{red}First}, {\color{blue}Second}, {\color{violet}Third}. Scores in bold are the best augmentation per method in each column.}
    \resizebox{\textwidth}{!}{
    \begin{tabular}{lllllllll}
    \toprule
      & USAir & NS & PB & Yeast & Celegans & Power & Router & Ecoli \\
    \midrule
    GCN (sup) & 89.74$\pm$2.29 & 80.91$\pm$1.18 & \color{red}{50.94$\pm$3.3} & 84.19$\pm$0.94 & 63.61$\pm$3.65 & 30.37$\pm$0.86 & 26.36$\pm$1.52 & 81.91$\pm$1.1 \\
    \midrule
    GRACE random & \color{red}{\textbf{91.48$\pm$1.72}} & 79.84$\pm$0.86 & 48.03$\pm$2.32 & 84.23$\pm$0.52 & \color{blue}{72.1$\pm$2.43} & 32.15$\pm$1.25 & 29.91$\pm$2.21 & 81.93$\pm$0.63 \\
    GRACE deg & 87.39$\pm$10.03 & 80.86$\pm$2.18 & 46.12$\pm$3.78 & 83.14$\pm$1.26 & 69.23$\pm$2.68 & 27.02$\pm$9.75 & 17.79$\pm$10.89 & 71.05$\pm$24.03 \\
    GRACE evc & 59.46$\pm$4.74 & 74.01$\pm$3.76 & 44.36$\pm$3.17 & 82.72$\pm$0.9 & 70.4$\pm$2.77 & 31.93$\pm$1.33 & 21.46$\pm$4.59 & 82.19$\pm$1.2 \\
    GRACE pr & \color{blue}{90.33$\pm$1.95} & 82.24$\pm$1.95 & \color{blue}{\textbf{49.56$\pm$3.68}} & 82.48$\pm$0.72 & \color{red}{\textbf{72.75$\pm$1.57}} & 33.45$\pm$1.24 & 53.76$\pm$2.91 & \color{red}{\textbf{84.32$\pm$0.85}} \\
    GRACE scom & 88.4$\pm$2.93 & 80.31$\pm$1.34 & 46.01$\pm$3.11 & 81.88$\pm$1.23 & 67.32$\pm$2.46 & 36.15$\pm$1.51 & 42.33$\pm$1.6 & 82.88$\pm$1.39 \\
    GRACE sbm oracle & 83.01$\pm$3.3 & 80.44$\pm$5.17 & 43.99$\pm$2.84 & \textbf{85.52$\pm$2.08} & 65.97$\pm$3.24 & 95.49$\pm$1.25 & \color{violet}{\textbf{96.63$\pm$0.94}} & 81.86$\pm$1.51 \\
    GRACE sbm2 oracle & 83.53$\pm$4.55 & \color{red}{\textbf{95.69$\pm$0.89}} & 37.12$\pm$2.94 & 85.28$\pm$0.9 & 62.17$\pm$5.04 & \color{violet}{\textbf{96.46$\pm$1.87}} & 94.7$\pm$2.67 & 81.25$\pm$1.35 \\
    \midrule
    CSGCL random & 86.26$\pm$3.28 & 81.73$\pm$0.87 & 47.92$\pm$2.02 & 81.68$\pm$1.07 & 70.09$\pm$2.1 & 35.08$\pm$1.32 & 40.11$\pm$2.23 & 81.5$\pm$1.11 \\
    CSGCL deg & 88.94$\pm$2.81 & 80.33$\pm$3.89 & 48.95$\pm$2.73 & 82.06$\pm$1.56 & \color{violet}{\textbf{71.7$\pm$2.12}} & 33.47$\pm$2.17 & 21.19$\pm$8.68 & 82.06$\pm$0.61 \\
    CSGCL evc & 81.53$\pm$2.98 & 71.33$\pm$2.75 & \color{violet}{\textbf{49.42$\pm$2.8}} & 83.53$\pm$0.8 & 70.96$\pm$3.13 & 35.12$\pm$1.39 & 27.19$\pm$1.47 & \color{blue}{\textbf{84.16$\pm$1.38}} \\
    CSGCL pr & 85.25$\pm$2.51 & 79.07$\pm$2.65 & 41.48$\pm$2.74 & 80.76$\pm$0.88 & 71.4$\pm$2.43 & 34.41$\pm$1.0 & 33.57$\pm$5.53 & 82.09$\pm$1.32 \\
    CSGCL scom & \color{violet}{\textbf{90.16$\pm$1.61}} & 82.45$\pm$1.69 & 47.75$\pm$3.5 & 83.32$\pm$0.72 & 69.37$\pm$3.29 & 34.58$\pm$1.1 & 45.59$\pm$3.15 & 74.07$\pm$1.43 \\
    CSGCL sbm oracle & 85.6$\pm$3.44 & \color{blue}{\textbf{95.27$\pm$0.62}} & 45.72$\pm$3.05 & \color{red}{\textbf{89.69$\pm$1.12}} & 64.41$\pm$2.79 & 96.35$\pm$0.88 & \color{blue}{97.42$\pm$0.7} & 78.67$\pm$1.49 \\
    CSGCL sbm2 oracle & 87.91$\pm$3.02 & 94.78$\pm$0.75 & 44.63$\pm$3.25 & 82.66$\pm$2.02 & 58.28$\pm$5.7 & \color{red}{\textbf{96.63$\pm$0.96}} & \color{red}{\textbf{97.55$\pm$0.97}} & 80.91$\pm$1.17 \\
    \midrule
    BGRL random & 87.11$\pm$1.9 & 80.49$\pm$1.29 & 47.24$\pm$3.24 & 82.72$\pm$1.01 & 67.41$\pm$3.44 & 25.67$\pm$0.74 & 30.95$\pm$1.16 & 81.63$\pm$0.72 \\
    BGRL deg & 86.47$\pm$3.22 & 83.47$\pm$2.74 & 47.75$\pm$3.57 & 80.99$\pm$1.49 & 68.72$\pm$3.37 & 28.92$\pm$2.63 & 41.41$\pm$6.38 & 82.73$\pm$0.83 \\
    BGRL evc & 88.82$\pm$1.47 & 80.42$\pm$0.81 & \textbf{49.0$\pm$3.9} & 84.25$\pm$0.57 & 60.47$\pm$4.94 & 21.4$\pm$1.2 & 47.79$\pm$6.0 & 82.79$\pm$1.45 \\
    BGRL pr & 86.24$\pm$2.51 & 80.29$\pm$1.4 & 44.47$\pm$2.73 & 81.65$\pm$1.54 & 66.67$\pm$3.77 & 26.05$\pm$1.12 & 48.2$\pm$2.92 & \color{violet}{\textbf{82.95$\pm$0.81}} \\
    BGRL scom & \textbf{89.74$\pm$2.13} & 81.3$\pm$0.8 & 44.67$\pm$2.61 & 84.08$\pm$0.74 & \textbf{70.51$\pm$3.43} & 25.86$\pm$1.07 & 54.25$\pm$3.26 & 82.36$\pm$1.04 \\
    BGRL sbm oracle & 87.76$\pm$2.86 & 94.65$\pm$1.1 & 43.83$\pm$3.02 & \color{violet}{86.74$\pm$1.48} & 55.8$\pm$3.38 & 94.77$\pm$2.28 & \textbf{87.87$\pm$1.94} & 81.65$\pm$0.69 \\
    BGRL sbm2 oracle & 87.46$\pm$2.54 & \color{violet}{\textbf{95.09$\pm$0.9}} & 45.0$\pm$2.84 & \color{blue}{\textbf{86.94$\pm$1.4}} & 59.09$\pm$3.19 & \color{blue}{\textbf{96.52$\pm$0.81}} & 75.97$\pm$2.35 & 82.25$\pm$0.75 \\
    \bottomrule
    \end{tabular}
    }
    \label{tab:sbm_oracle}
\end{table*}
The Oracle version of the SBM augmentation achieves the best score with a significant margin on the NS, Yeast, Power, and Router datasets. 
The highest scores are obtained on Power and Router, probably because they are infrastructure graphs, in which the presence of links between nodes is dependent on the geographical distance between them, which can be approximated with the community structure. Nevertheless, the oracle version doesn't work on all graphs; for example, it works nicely on NS but fails to reach good performance on the PB graph, probably because local information is more important than community structure in this graph. These results show that if we are indeed able to detect good communities, and if the community structure is relevant for a dataset, then the SBM augmentation can bring a major gain in link prediction.

\begin{table*}[h!t]
    \centering
    \caption{Link prediction results with Hits@50 metric, of the sbm augmentation, on non-attributed graphs. Supervised methods are reported (sup). The top three scores per column are colored by {\color{red}First}, {\color{blue}Second}, {\color{violet}Third}. Scores in bold are the best augmentation per method in each column.}
    \resizebox{\textwidth}{!}{
    \begin{tabular}{lllllllll}
    \toprule
     & USAir & NS & PB & Yeast & Celegans & Power & Router & Ecoli \\
    \midrule
    GCN (sup) & 89.74$\pm$2.29 & 80.91$\pm$1.18 & \color{red}{50.94$\pm$3.3} & \color{violet}{84.19$\pm$0.94} & 63.61$\pm$3.65 & 30.37$\pm$0.86 & 26.36$\pm$1.52 & 81.91$\pm$1.1 \\
    \midrule
    GRACE random & \color{red}{\textbf{91.48$\pm$1.72}} & 79.84$\pm$0.86 & 48.03$\pm$2.32 & \color{blue}{\textbf{84.23$\pm$0.52}} & \color{blue}{72.1$\pm$2.43} & 32.15$\pm$1.25 & 29.91$\pm$2.21 & 81.93$\pm$0.63 \\
    GRACE deg & 87.39$\pm$10.03 & 80.86$\pm$2.18 & 46.12$\pm$3.78 & 83.14$\pm$1.26 & 69.23$\pm$2.68 & 27.02$\pm$9.75 & 17.79$\pm$10.89 & 71.05$\pm$24.03 \\
    GRACE evc & 59.46$\pm$4.74 & 74.01$\pm$3.76 & 44.36$\pm$3.17 & 82.72$\pm$0.9 & 70.4$\pm$2.77 & 31.93$\pm$1.33 & 21.46$\pm$4.59 & 82.19$\pm$1.2 \\
    GRACE pr & \color{blue}{90.33$\pm$1.95} & \color{violet}{\textbf{82.24$\pm$1.95}} & \color{blue}{\textbf{49.56$\pm$3.68}} & 82.48$\pm$0.72 & \color{red}{\textbf{72.75$\pm$1.57}} & 33.45$\pm$1.24 & \color{blue}{\textbf{53.76$\pm$2.91}} & \color{red}{\textbf{84.32$\pm$0.85}} \\
    GRACE scom & 88.4$\pm$2.93 & 80.31$\pm$1.34 & 46.01$\pm$3.11 & 81.88$\pm$1.23 & 67.32$\pm$2.46 & \color{blue}{\textbf{36.15$\pm$1.51}} & 42.33$\pm$1.6 & 82.88$\pm$1.39 \\
    GRACE sbm & 87.01$\pm$2.03 & 73.72$\pm$4.33 & 45.65$\pm$2.74 & 79.84$\pm$1.61 & 64.71$\pm$3.17 & 32.61$\pm$1.17 & 33.83$\pm$1.93 & 79.54$\pm$1.14 \\
    GRACE sbm2 & 86.21$\pm$3.63 & 80.89$\pm$1.41 & 45.54$\pm$2.5 & 79.46$\pm$1.49 & 61.98$\pm$3.55 & 33.87$\pm$1.68 & 26.43$\pm$2.69 & 82.36$\pm$0.64 \\
    \midrule
    CSGCL random & 86.26$\pm$3.28 & 81.73$\pm$0.87 & 47.92$\pm$2.02 & 81.68$\pm$1.07 & 70.09$\pm$2.1 & 35.08$\pm$1.32 & 40.11$\pm$2.23 & 81.5$\pm$1.11 \\
    CSGCL deg & 88.94$\pm$2.81 & 80.33$\pm$3.89 & 48.95$\pm$2.73 & 82.06$\pm$1.56 & \color{violet}{\textbf{71.7$\pm$2.12}} & 33.47$\pm$2.17 & 21.19$\pm$8.68 & 82.06$\pm$0.61 \\
    CSGCL evc & 81.53$\pm$2.98 & 71.33$\pm$2.75 & \color{violet}{\textbf{49.42$\pm$2.8}} & \textbf{83.53$\pm$0.8} & 70.96$\pm$3.13 & 35.12$\pm$1.39 & 27.19$\pm$1.47 & \color{blue}{\textbf{84.16$\pm$1.38}} \\
    CSGCL pr & 85.25$\pm$2.51 & 79.07$\pm$2.65 & 41.48$\pm$2.74 & 80.76$\pm$0.88 & 71.4$\pm$2.43 & 34.41$\pm$1.0 & 33.57$\pm$5.53 & 82.09$\pm$1.32 \\
    CSGCL scom & \color{violet}{\textbf{90.16$\pm$1.61}} & \color{blue}{\textbf{82.45$\pm$1.69}} & 47.75$\pm$3.5 & 83.32$\pm$0.72 & 69.37$\pm$3.29 & 34.58$\pm$1.1 & \color{violet}{\textbf{45.59$\pm$3.15}} & 74.07$\pm$1.43 \\
    CSGCL sbm & 86.75$\pm$2.73 & 80.51$\pm$1.09 & 43.37$\pm$3.21 & 81.51$\pm$1.29 & 51.59$\pm$2.58 & 35.14$\pm$1.83 & 27.74$\pm$1.49 & 75.1$\pm$0.82 \\
    CSGCL sbm2 & 88.89$\pm$2.39 & 80.2$\pm$1.22 & 44.97$\pm$2.46 & 79.95$\pm$1.41 & 54.03$\pm$1.83 & \color{red}{\textbf{36.22$\pm$1.6}} & 27.03$\pm$3.64 & 80.45$\pm$1.06 \\
    \midrule
    BGRL random & 87.11$\pm$1.9 & 80.49$\pm$1.29 & 47.24$\pm$3.24 & 82.72$\pm$1.01 & 67.41$\pm$3.44 & 25.67$\pm$0.74 & 30.95$\pm$1.16 & 81.63$\pm$0.72 \\
    BGRL deg & 86.47$\pm$3.22 & \color{red}{\textbf{83.47$\pm$2.74}} & 47.75$\pm$3.57 & 80.99$\pm$1.49 & 68.72$\pm$3.37 & 28.92$\pm$2.63 & 41.41$\pm$6.38 & 82.73$\pm$0.83 \\
    BGRL evc & 88.82$\pm$1.47 & 80.42$\pm$0.81 & \textbf{49.0$\pm$3.9} & \color{red}{\textbf{84.25$\pm$0.57}} & 60.47$\pm$4.94 & 21.4$\pm$1.2 & 47.79$\pm$6.0 & 82.79$\pm$1.45 \\
    BGRL pr & 86.24$\pm$2.51 & 80.29$\pm$1.4 & 44.47$\pm$2.73 & 81.65$\pm$1.54 & 66.67$\pm$3.77 & 26.05$\pm$1.12 & 48.2$\pm$2.92 & \color{violet}{\textbf{82.95$\pm$0.81}} \\
    BGRL scom & \textbf{89.74$\pm$2.13} & 81.3$\pm$0.8 & 44.67$\pm$2.61 & 84.08$\pm$0.74 & \textbf{70.51$\pm$3.43} & 25.86$\pm$1.07 & \color{red}{\textbf{54.25$\pm$3.26}} & 82.36$\pm$1.04 \\
    BGRL sbm & 88.82$\pm$2.35 & 79.18$\pm$1.49 & 43.95$\pm$2.45 & 81.01$\pm$1.37 & 58.72$\pm$3.18 & \color{violet}{\textbf{35.34$\pm$1.33}} & 49.48$\pm$3.53 & 79.21$\pm$1.4 \\
    BGRL sbm2 & 80.59$\pm$3.58 & 80.46$\pm$1.36 & 42.7$\pm$2.91 & 79.58$\pm$1.59 & 57.2$\pm$3.6 & 31.18$\pm$1.64 & 43.33$\pm$3.33 & 79.77$\pm$1.02 \\
    \bottomrule
    \end{tabular}
    }
    \label{tab:sbm}
\end{table*}
In Table \ref{tab:sbm}, we present sbm augmentation with communities detected only on the train split, as expected in real settings. 
As expected, these SBM augmentations achieve scores lower or equal than the SBM oracle. But even though the oracle version reaches high scores on four datasets(NS, Yeast, Power, and Router), the sbm augmentation does not improve over baselines on three of them (NS, Yeast, and Router). We can explain this difference between with and without the Oracle by the random splitting; not enough edges remain in the training edges set to retrieve the real community structure of the Graph. On the Power Graph, for the three models GRACE, CSGCL, and BGRL, the sbm augmentation achieves high scores, respectively 33.87$\pm$1.68, 36.22$\pm$1.6, and 35.34$\pm$1.33. Comparing the tree models, CSGCL achieves the highest scores, irrespective of the augmentation choice, showing that its loss tuning in favor of integrating community information works well in this Graph. We can deduce that the Power graph structure relies strongly on the community structure, even with the ablation of 30\% of edges (random splitting). Also, the community detection is good enough in this graph to bring structural information through the augmentation process.

If we focus on comparing augmentation, we observe that the performance depends strongly on the augmentation process. In fact, the random augmentation is always outperformed by another augmentation for the BGRL model, as opposed to the results on node classification obtained by the authors of BGRL. We can make the same observation on the CSGCL model, and the majority of the time on the GRACE model, with 6 times out of 8 where GRACE random is worse than another adaptive augmentation. Moreover, most of the time, the gap between the worst and the best augmentation choice is large (eg. on Router Graph: GRACE deg score 17.79$\pm$10.89 and GRACE pr score 53.76$\pm$2.91).

We can extract some tendencies of the augmentation process and the model combination, with for example, GRACE and the adaptive augmentation pr (page rank) which is the best augmentation for the GRACE model on 5 datasets and in the top tree on 6 datasets. In the same way, we can find negative correlation like GRACE deg, which on 7 datasets performs poorly against the random augmentation. We observe that on CSGCL, the scom augmentation proposed in the original article is not always the best (3 out of 8), and it shows that the model itself can benefit from varying adaptive augmentation depending on the input graph. Thus, the correlation between the augmentation and the dataset exists at some point. As said, the community augmentation scom or sbm and sbm2 seem appropriate on the Power Dataset. On Yeast, the EVC augmentation achieves pretty good scores on the three models. 

From this analysis, we see that both the augmentation choice and the model have important effects, but that their performance depends strongly on the dataset. 
We thus argue that the augmentation process has to be treated as a hyperparameter of the model, which is important to tune, because none of them is overall better on all datasets and all models.
In the rest of the paper, we thus present our results using an "optim" category, corresponding to the maximum score of adaptive augmentations for each model and graph. 

\subsubsection{L-GRACE and L-BGRL vs GRACE and BGRL}

To evaluate the interest of our proposed architectures/models L-GRACE and L-BGRL (see section \ref{adapted model}), we compare them to MPLP reported scores, on the 8 non-attributed graphs. Table \ref{tab:non-attr_simple} shows full results including L-GRACE and L-BGRL in simplified form with the aforesaid "optim" abbreviation, which corresponds to \textit{the best adaptive augmentation choice for a model and a graph}. The Table \ref{tab:non-attr} is the details forms with each augmentation per model; in addition, the reader can refer to the table \ref{tab:non-attr AP} and table \ref{tab:non-attr AUC} in the Appendix section \ref{appendix} for AP and ROC-AUC score.
\begin{table*}[h!t]
    \centering
    \caption{Simplified link prediction results with Hits@50 metric, on non-attributed graphs. Reported methods from the MPLP article are noted with *. Supervised methods are reported (sup). The top three models are colored by {\color{red}First}, {\color{blue}Second}, {\color{violet}Third}. The best scores from all SSL models are underlined.}
    \resizebox{\textwidth}{!}{
    \begin{tabular}{lllllllll}
    \toprule
     & USAir & NS & PB & Yeast & Celegans & Power & Router & Ecoli \\
    \midrule
    GCN* (sup) & 73.29$\pm$4.70 & 78.32$\pm$2.57 & 37.32$\pm$4.69 & 73.15$\pm$2.41 & 40.68$\pm$5.45 & 15.40$\pm$2.90 & 24.42$\pm$4.59 & 61.02$\pm$11.91 \\
    SEAL* (sup) & 90.47$\pm$3.00 & 86.59$\pm$3.03 & 44.47$\pm$2.86 & 83.92$\pm$1.17 & 64.80$\pm$4.23 & 31.46$\pm$3.25 & \color{violet}{61.00$\pm$10.10} & 83.42$\pm$1.01 \\
    ELPH* (sup) & 87.60$\pm$1.49 & \color{violet}{88.49$\pm$2.14} & 46.91$\pm$2.21 & 82.74$\pm$1.19 & 64.45$\pm$3.91 & 26.61$\pm$1.73 & \color{blue}{61.07$\pm$3.06} & 75.25$\pm$1.44 \\
    NCNC* (sup) & 86.16$\pm$1.77 & 83.18$\pm$3.17 & 46.85$\pm$3.18 & 82.00$\pm$0.97 & 60.49$\pm$5.09 & 23.28$\pm$1.55 & 52.45$\pm$8.77 & 83.94$\pm$1.57 \\
    MPLP* (sup) & \color{blue}{92.12$\pm$2.21} & \color{red}{90.02$\pm$2.04} & \color{red}{52.55$\pm$2.90} & \color{red}{85.36$\pm$0.72} & \color{red}{74.28$\pm$2.09} & 32.66$\pm$3.58 & \color{red}{64.68$\pm$3.14} & \color{blue}{86.11$\pm$0.83} \\
    MPLP+* (sup) & 91.24$\pm$2.11 & \color{blue}{88.91$\pm$2.04} & \color{violet}{51.81$\pm$2.39} & \color{blue}{84.95$\pm$0.66} & 72.73$\pm$2.99 & 31.86$\pm$2.59 & 60.94$\pm$2.51 & \color{red}{87.07$\pm$0.89} \\
    \midrule
    GCN (sup) & 89.74$\pm$2.29 & 80.91$\pm$1.18 & 50.94$\pm$3.3 & 84.19$\pm$0.94 & 63.61$\pm$3.65 & 30.37$\pm$0.86 & 26.36$\pm$1.52 & 81.91$\pm$1.1 \\
    \midrule
    GRACE random & \color{violet}{91.48$\pm$1.72} & 79.84$\pm$0.86 & 48.03$\pm$2.32 & 84.23$\pm$0.52 & 72.1$\pm$2.43 & 32.15$\pm$1.25 & 29.91$\pm$2.21 & 81.93$\pm$0.63 \\
    GRACE optim & 90.33$\pm$1.95 & 82.24$\pm$1.95 & 49.56$\pm$3.68 & 83.14$\pm$1.26 & \color{violet}{72.75$\pm$1.57} & \color{blue}{36.15$\pm$1.51} & 53.76$\pm$2.91 & \color{violet}{\underline{84.32$\pm$0.85}} \\
    \midrule
    CSGCL random & 86.26$\pm$3.28 & 81.73$\pm$0.87 & 47.92$\pm$2.02 & 81.68$\pm$1.07 & 70.09$\pm$2.1 & 35.08$\pm$1.32 & 40.11$\pm$2.23 & 81.5$\pm$1.11 \\
    CSGCL optim & 90.16$\pm$1.61 & 82.45$\pm$1.69 & 49.42$\pm$2.8 & 83.53$\pm$0.8 & 71.7$\pm$2.12 & \color{red}{\underline{36.22$\pm$1.6}} & 45.59$\pm$3.15 & 84.16$\pm$1.38 \\
    \midrule
    BGRL random & 87.11$\pm$1.9 & 80.49$\pm$1.29 & 47.24$\pm$3.24 & 82.72$\pm$1.01 & 67.41$\pm$3.44 & 25.67$\pm$0.74 & 30.95$\pm$1.16 & 81.63$\pm$0.72 \\
    BGRL optim & 89.74$\pm$2.13 & 83.47$\pm$2.74 & 49.0$\pm$3.9 & 84.25$\pm$0.57 & 70.51$\pm$3.43 & \color{violet}{35.34$\pm$1.33} & \underline{54.25$\pm$3.26} & 82.95$\pm$0.81 \\
    \midrule
    L-GRACE random & \color{red}{\underline{92.47$\pm$0.8}} & 80.8$\pm$1.45 & \color{blue}{\underline{51.98$\pm$3.28}} & 84.33$\pm$1.09 & 70.56$\pm$3.27 & 30.58$\pm$1.81 & 34.67$\pm$10.92 & 81.67$\pm$1.22 \\
    L-GRACE optim & 90.8$\pm$1.84 & 83.34$\pm$1.49 & 50.51$\pm$3.81 & \color{violet}{\underline{84.38$\pm$0.68}} & \color{blue}{\underline{72.89$\pm$2.3}} & 35.08$\pm$1.91 & 29.76$\pm$3.38 & 84.27$\pm$0.78 \\
    \midrule
    L-BGRL random & 79.81$\pm$3.92 & 80.02$\pm$1.24 & 47.72$\pm$2.3 & 81.52$\pm$1.16 & 65.52$\pm$2.25 & 23.66$\pm$0.95 & 39.72$\pm$8.46 & 81.75$\pm$1.01 \\
    L-BGRL optim & 90.66$\pm$1.93 & \underline{85.0$\pm$1.88} & 47.68$\pm$3.35 & 84.22$\pm$0.73 & 71.03$\pm$4.36 & 34.94$\pm$1.37 & 49.9$\pm$3.3 & 83.34$\pm$1.08 \\
    \bottomrule
    \end{tabular}
    }
    \label{tab:non-attr_simple}
\end{table*}
\begin{table*}[h!t]
    \centering
    \caption{Full link prediction results with the Hits@50 metric, on non-attributed graphs. Reported methods from the MPLP article are noted with *. Supervised methods are reported (sup). The top three models are colored by {\color{red}First}, {\color{blue}Second}, {\color{violet}Third}. Scores in bold are the best augmentation per method in each column. $\bigstar$ shows the most significant best results, and X shows the most significant worst results from our experimentation. The best scores from all SSL models are underlined.}
    \resizebox{\textwidth}{!}{
    \begin{tabular}{lllllllll}
    \toprule
     & USAir & NS & PB & Yeast & Celegans & Power & Router & Ecoli \\
    \midrule
    CN* & 80.52$\pm$4.07 & 74$\pm$1.98 & 37.22$\pm$3.52 & 72.6$\pm$3.85 & 47.67$\pm$10.87 & 11.57$\pm$0.55 & 9.38$\pm$1.05 & 51.74$\pm$2.7 \\
    AA* & 85.51$\pm$2.25 & 74$\pm$1.98 & 39.48$\pm$3.53 & 73.62$\pm$1.01 & 58.34$\pm$2.88 & 11.57$\pm$0.55 & 9.38$\pm$1.05 & 68.13$\pm$1.61 \\
    RA* & 85.95$\pm$1.83 & 74$\pm$1.98 & 39.84$\pm$3.54 & 73.62$\pm$1.01 & 61.47$\pm$4.59 & 11.57$\pm$0.55 & 9.38$\pm$1.05 & 74.45$\pm$0.55 \\
    GCN* (sup) & 73.29$\pm$4.70 & 78.32$\pm$2.57 & 37.32$\pm$4.69 & 73.15$\pm$2.41 & 40.68$\pm$5.45 & 15.40$\pm$2.90 & 24.42$\pm$4.59 & 61.02$\pm$11.91 \\
    SEAL* (sup) & 90.47$\pm$3.00 & 86.59$\pm$3.03 & 44.47$\pm$2.86 & 83.92$\pm$1.17 & 64.80$\pm$4.23 & 31.46$\pm$3.25 & \color{violet}{61.00$\pm$10.10} & 83.42$\pm$1.01 \\
    ELPH* (sup) & 87.60$\pm$1.49 & \color{violet}{88.49$\pm$2.14} & 46.91$\pm$2.21 & 82.74$\pm$1.19 & 64.45$\pm$3.91 & 26.61$\pm$1.73 & \color{blue}{61.07$\pm$3.06} & 75.25$\pm$1.44 \\
    NCNC* (sup) & 86.16$\pm$1.77 & 83.18$\pm$3.17 & 46.85$\pm$3.18 & 82.00$\pm$0.97 & 60.49$\pm$5.09 & 23.28$\pm$1.55 & 52.45$\pm$8.77 & 83.94$\pm$1.57 \\
    MPLP* (sup) & \color{blue}{92.12$\pm$2.21} & \color{red}{90.02$\pm$2.04} & \color{red}{52.55$\pm$2.90} & \color{red}{85.36$\pm$0.72} & \color{red}{74.28$\pm$2.09} & 32.66$\pm$3.58 & \color{red}{64.68$\pm$3.14} & \color{blue}{86.11$\pm$0.83} \\
    MPLP+* (sup) & 91.24$\pm$2.11 & \color{blue}{88.91$\pm$2.04} & \color{violet}{51.81$\pm$2.39} & \color{blue}{84.95$\pm$0.66} & 72.73$\pm$2.99 & 31.86$\pm$2.59 & 60.94$\pm$2.51 & \color{red}{87.07$\pm$0.89} \\
    \midrule
    GCN (sup) & 89.74$\pm$2.29 & 80.91$\pm$1.18 & 50.94$\pm$3.3 & 84.19$\pm$0.94 $\bigstar$ & 63.61$\pm$3.65 & 30.37$\pm$0.86 & 26.36$\pm$1.52 & 81.91$\pm$1.1 \\
    \midrule
    GRACE random & \color{violet}{\textbf{91.48$\pm$1.72}} & 79.84$\pm$0.86 & 48.03$\pm$2.32 & \textbf{84.23$\pm$0.52 $\bigstar$} & 72.1$\pm$2.43 & 32.15$\pm$1.25 & 29.91$\pm$2.21 & 81.93$\pm$0.63 \\
    GRACE deg & 87.39$\pm$10.03 & 80.86$\pm$2.18 & 46.12$\pm$3.78 & 83.14$\pm$1.26 & 69.23$\pm$2.68 & 27.02$\pm$9.75 & 17.79$\pm$10.89 X & 71.05$\pm$24.03 \\
    GRACE evc & 59.46$\pm$4.74 X & 74.01$\pm$3.76 & 44.36$\pm$3.17 & 82.72$\pm$0.9 & 70.4$\pm$2.77 & 31.93$\pm$1.33 & 21.46$\pm$4.59 X & 82.19$\pm$1.2 \\
    GRACE pr & 90.33$\pm$1.95 & \textbf{82.24$\pm$1.95} & \textbf{49.56$\pm$3.68} & 82.48$\pm$0.72 & \color{violet}{\textbf{72.75$\pm$1.57 $\bigstar$}} & 33.45$\pm$1.24 & \textbf{53.76$\pm$2.91 $\bigstar$} & \color{violet}{\textbf{\underline{84.32$\pm$0.85 $\bigstar$}}} \\
    GRACE scom & 88.4$\pm$2.93 & 80.31$\pm$1.34 & 46.01$\pm$3.11 & 81.88$\pm$1.23 & 67.32$\pm$2.46 & \color{blue}{\textbf{36.15$\pm$1.51 $\bigstar$}} & 42.33$\pm$1.6 & 82.88$\pm$1.39 \\
    GRACE sbm & 87.01$\pm$2.03 & 73.72$\pm$4.33 & 45.65$\pm$2.74 & 79.84$\pm$1.61 X & 64.71$\pm$3.17 & 32.61$\pm$1.17 & 33.83$\pm$1.93 & 79.54$\pm$1.14 \\
    GRACE sbm2 & 86.21$\pm$3.63 & 80.89$\pm$1.41 & 45.54$\pm$2.5 & 79.46$\pm$1.49 X & 61.98$\pm$3.55 & 33.87$\pm$1.68 & 26.43$\pm$2.69 & 82.36$\pm$0.64 \\
    \midrule
    CSGCL random & 86.26$\pm$3.28 & 81.73$\pm$0.87 & 47.92$\pm$2.02 & 81.68$\pm$1.07 & 70.09$\pm$2.1 & 35.08$\pm$1.32 & 40.11$\pm$2.23 & 81.5$\pm$1.11 \\
    CSGCL deg & 88.94$\pm$2.81 & 80.33$\pm$3.89 & 48.95$\pm$2.73 & 82.06$\pm$1.56 & \textbf{71.7$\pm$2.12} & 33.47$\pm$2.17 & 21.19$\pm$8.68 & 82.06$\pm$0.61 \\
    CSGCL evc & 81.53$\pm$2.98 & 71.33$\pm$2.75 & \textbf{49.42$\pm$2.8} & \textbf{83.53$\pm$0.8} & 70.96$\pm$3.13 & 35.12$\pm$1.39 & 27.19$\pm$1.47 & \textbf{84.16$\pm$1.38} \\
    CSGCL pr & 85.25$\pm$2.51 & 79.07$\pm$2.65 & 41.48$\pm$2.74 & 80.76$\pm$0.88 & 71.4$\pm$2.43 & 34.41$\pm$1.0 & 33.57$\pm$5.53 & 82.09$\pm$1.32 \\
    CSGCL scom & \textbf{90.16$\pm$1.61} & \textbf{82.45$\pm$1.69} & 47.75$\pm$3.5 & 83.32$\pm$0.72 & 69.37$\pm$3.29 & 34.58$\pm$1.1 & \textbf{45.59$\pm$3.15} & 74.07$\pm$1.43 X \\
    CSGCL sbm & 86.75$\pm$2.73 & 80.51$\pm$1.09 & 43.37$\pm$3.21 & 81.51$\pm$1.29 & 51.59$\pm$2.58 X & 35.14$\pm$1.83 & 27.74$\pm$1.49 & 75.1$\pm$0.82 X \\
    CSGCL sbm2 & 88.89$\pm$2.39 & 80.2$\pm$1.22 & 44.97$\pm$2.46 & 79.95$\pm$1.41 & 54.03$\pm$1.83 & \color{red}{\textbf{\underline{36.22$\pm$1.6 $\bigstar$}}} & 27.03$\pm$3.64 & 80.45$\pm$1.06 \\
    \midrule
    BGRL random & 87.11$\pm$1.9 & 80.49$\pm$1.29 & 47.24$\pm$3.24 & 82.72$\pm$1.01 & 67.41$\pm$3.44 & 25.67$\pm$0.74 & 30.95$\pm$1.16 & 81.63$\pm$0.72 \\
    BGRL deg & 86.47$\pm$3.22 & \textbf{83.47$\pm$2.74} & 47.75$\pm$3.57 & 80.99$\pm$1.49 & 68.72$\pm$3.37 & 28.92$\pm$2.63 & 41.41$\pm$6.38 & 82.73$\pm$0.83 \\
    BGRL evc & 88.82$\pm$1.47 & 80.42$\pm$0.81 & \textbf{49.0$\pm$3.9} & \textbf{84.25$\pm$0.57 $\bigstar$} & 60.47$\pm$4.94 & 21.4$\pm$1.2 X & 47.79$\pm$6.0 & 82.79$\pm$1.45 \\
    BGRL pr & 86.24$\pm$2.51 & 80.29$\pm$1.4 & 44.47$\pm$2.73 & 81.65$\pm$1.54 & 66.67$\pm$3.77 & 26.05$\pm$1.12 & 48.2$\pm$2.92 & \textbf{82.95$\pm$0.81} \\
    BGRL scom & \textbf{89.74$\pm$2.13} & 81.3$\pm$0.8 & 44.67$\pm$2.61 & 84.08$\pm$0.74 & \textbf{70.51$\pm$3.43} & 25.86$\pm$1.07 & \textbf{\underline{54.25$\pm$3.26 $\bigstar$}} & 82.36$\pm$1.04 \\
    BGRL sbm & 88.82$\pm$2.35 & 79.18$\pm$1.49 & 43.95$\pm$2.45 & 81.01$\pm$1.37 & 58.72$\pm$3.18 & \color{violet}{\textbf{35.34$\pm$1.33}} & 49.48$\pm$3.53 & 79.21$\pm$1.4 \\
    BGRL sbm2 & 80.59$\pm$3.58 & 80.46$\pm$1.36 & 42.7$\pm$2.91 & 79.58$\pm$1.59 X & 57.2$\pm$3.6 & 31.18$\pm$1.64 & 43.33$\pm$3.33 & 79.77$\pm$1.02 \\
    \midrule
    L-GRACE random & \color{red}{\textbf{\underline{92.47$\pm$0.8 $\bigstar$}}} & 80.8$\pm$1.45 & \color{blue}{\textbf{\underline{51.98$\pm$3.28 $\bigstar$}}} & 84.33$\pm$1.09 & 70.56$\pm$3.27 & 30.58$\pm$1.81 & \textbf{34.67$\pm$10.92} & 81.67$\pm$1.22 \\
    L-GRACE deg & 90.71$\pm$1.88 & 80.27$\pm$1.44 & 48.52$\pm$2.79 & 82.03$\pm$1.59 & 65.9$\pm$3.9 & 33.7$\pm$2.51 & 29.34$\pm$6.84 & \textbf{84.27$\pm$0.78 $\bigstar$} \\
    L-GRACE evc & 90.61$\pm$2.57 & 83.16$\pm$1.43 & 50.51$\pm$3.81 & 83.75$\pm$1.05 & 72.54$\pm$4.24 $\bigstar$ & 28.83$\pm$0.89 & 26.09$\pm$5.79 & 79.19$\pm$0.93 \\
    L-GRACE pr & 90.0$\pm$2.02 & \textbf{83.34$\pm$1.49} & 48.7$\pm$2.61 & \color{violet}{\textbf{\underline{84.38$\pm$0.68 $\bigstar$}}} & \color{blue}{\textbf{\underline{72.89$\pm$2.3 $\bigstar$}}} & 29.5$\pm$1.19 & 33.12$\pm$1.88 & 83.08$\pm$0.45 \\
    L-GRACE scom & 90.8$\pm$1.84 & 81.13$\pm$1.21 & 49.92$\pm$2.82 & 79.85$\pm$7.51 & 65.31$\pm$4.03 & 27.67$\pm$1.68 & 31.41$\pm$1.33 & 82.45$\pm$0.88 \\
    L-GRACE sbm & 90.38$\pm$2.54 & 53.41$\pm$17.6 X & 48.54$\pm$2.99 & 83.22$\pm$0.96 & 67.02$\pm$2.6 & 33.45$\pm$1.08 & 29.76$\pm$3.38 & 82.9$\pm$0.59 \\
    L-GRACE sbm2 & 88.33$\pm$2.57 & 80.58$\pm$1.14 & 44.43$\pm$4.54 & 79.77$\pm$1.78 & 60.44$\pm$3.16 & \textbf{35.08$\pm$1.91} & 27.6$\pm$1.06 & 81.28$\pm$1.0 \\
    \midrule
    L-BGRL random & 79.81$\pm$3.92 & 80.02$\pm$1.24 & \textbf{47.72$\pm$2.3} & 81.52$\pm$1.16 & 65.52$\pm$2.25 & 23.66$\pm$0.95 & 39.72$\pm$8.46 & 81.75$\pm$1.01 \\
    L-BGRL deg & 83.6$\pm$3.96 & 81.51$\pm$1.81 & 46.8$\pm$3.25 & 81.92$\pm$1.04 & \textbf{71.03$\pm$4.36} & 27.76$\pm$2.46 & \textbf{49.9$\pm$3.3} & 73.27$\pm$1.14 X \\
    L-BGRL evc & 89.39$\pm$2.4 & 67.96$\pm$4.61 X & 34.51$\pm$3.75 X & \textbf{84.22$\pm$0.73 $\bigstar$} & 67.3$\pm$2.55 & 27.54$\pm$0.92 & 34.83$\pm$4.65 & 82.07$\pm$0.93 \\
    L-BGRL pr & \textbf{90.66$\pm$1.93} & \textbf{\underline{85.0$\pm$1.88 $\bigstar$}} & 47.59$\pm$2.18 & 83.52$\pm$1.12 & 70.05$\pm$2.87 & 19.58$\pm$1.23 X & 45.02$\pm$6.75 & \textbf{83.34$\pm$1.08} \\
    L-BGRL scom & 89.22$\pm$2.76 & 76.86$\pm$1.38 & 47.68$\pm$3.35 & 83.14$\pm$0.55 & 69.04$\pm$2.79 & 20.4$\pm$0.92 X & 33.96$\pm$1.24 & 83.22$\pm$0.71 \\
    L-BGRL sbm & 79.06$\pm$2.75 & 84.31$\pm$1.97 $\bigstar$ & 44.46$\pm$2.91 & 80.77$\pm$1.18 & 65.22$\pm$2.76 & \textbf{34.94$\pm$1.37} & 42.29$\pm$3.98 & 77.81$\pm$0.98 \\
    L-BGRL sbm2 & 87.95$\pm$2.58 & 82.97$\pm$1.67 & 44.23$\pm$2.29 & 81.19$\pm$0.91 & 42.73$\pm$2.57 X & 33.11$\pm$1.47 & 35.94$\pm$2.44 & 79.21$\pm$1.47 \\
    \bottomrule
    \end{tabular}
    }
    \label{tab:non-attr}
\end{table*}
On the 8 chosen graphs, the best instance discrimination (SSL) always beats the classic supervised GCN. Particularly, with our proposed model L-GRACE, we obtain the best performance on USAir, and the second best for PB and Celegans, just behind the MPLP scores. Regarding model comparison, the link-prediction-specific loss function obtains the best performance among self-supervised models, dominating on five datasets out of eight. For the GRACE model, our improved loss permit to reach a higher score five times out of eight, showing that our link loss captures more information than the original one. On the BGRL model, the link-loss version L-BGRL improves the higher score on four datasets out of eight, mitigating our upgrade on asymmetric instance discrimination models, but on the NS dataset, L-BGRL is the best self-supervised model. Nevertheless, comparing the contrastive approach against the asymmetric one, we find that contrastive methods reach better performances than BGRL or L-BGRL globally.

Still comparing SSL models, in Table \ref{tab:non-attr}, the statistical test (Friedman test \cite{friedman1937use} followed by a Bonferroni-Dunn test \cite{dunn1961multiple}) informs the distributions of scores and indicates with a star $\bigstar$ the significantly best methods and with a cross X significant worst methods. Counting them, L-GRACE is 6 times in the group of the significant best methods and only one time in the group of worst, while GRACE is 5 times in the group of significant best, but 5 times in the group of worst. These results permits to admit that globally L-GRACE is less invariant to the augmentation process unlike other models, seeing that L-GRACE achieves most of the time results as good as the others, often best results and very rarely worse results, no matter the augmentation process.

Even if L-GRACE is less invariant to augmentation, we can observe some relationship between this model and some augmentations. The pr augmentation is the best for L-GRACE among the three datasets, and the random augmentation is too for the three other datasets. Of course, the sbm augmentation works great on L-GRACE and L-BGRL for the Power graph, but not as well as CSGCL sbm2, which performs the new SotA score for this graph. We can notice that the best augmentation per model is rarely the same between GRACE and its new version, L-GRACE.

\textbf{Extended evaluation on Attributed Graphs.} Even if this is not the main focus of our models, we also compare them on attributed graphs (see Table \ref{tab:attr-simple}), comparing with T-BGRL scores. For full comparison of each augmentation, see the table \ref{tab:attr} and for AP and ROC-AUC score, the reader can refer to the table \ref{tab:attr AP} and table \ref{tab:attr AUC} in the Appendix section \ref{appendix}. Our adaptation of loss and augmentation for link prediction was designed to use the structural context of the graph. We have chosen to focus our work on structural properties, hence the evaluation on non-attributed graphs. Only on Cora and Citeseer, all our instance discrimination models are better than the GCN supervised baseline, while on other datasets, the instance discrimination models failed to surpass it.

\begin{table*}[h!t]
    \centering
    \caption{Simplified link prediction results with Hits@50 metric, on attributed graphs. Reported methods from the T-BGRL article are noted with *. Supervised methods are reported (sup). The top three models are colored by {\color{red}First}, {\color{blue}Second}, {\color{violet}Third}. The optim scores from the SSL model are underlined.}
    \resizebox{\textwidth}{!}{
    \begin{tabular}{lllllll}
    \toprule
     & cora & citeseer & cs & physics & computers & photo \\
    \midrule
    GCN* (sup) & 81.6$\pm$1.3 & 82.2$\pm$1.17 & \color{blue}{76.2$\pm$1} & \color{blue}{79.8$\pm$1.8} & 42.6$\pm$3.6 & \color{violet}{64.2$\pm$2.9} \\
    GRACE random* & 68.6$\pm$5.6 & 70.7$\pm$6.8 & 45.6$\pm$6.6 & OOM & 24$\pm$2.7 & 48.6$\pm$2.5 \\
    BGRL random* & 79.2$\pm$1.5 & 85.8$\pm$2 & 51.5$\pm$1.6 & 47.6$\pm$1.5 & 34.6$\pm$1.8 & 56.2$\pm$1.3 \\
    T-BGRL random* & 77.3$\pm$2 & 86.8$\pm$2.3 & 55.5$\pm$0.9 & 47.1$\pm$2.1 & 31.5$\pm$1.5 & 51.7$\pm$1.6 \\
    \midrule
    GCN (sup) & 81.97$\pm$2.31 & 75.01$\pm$3.03 & \color{red}{83.89$\pm$1.28} & \color{red}{83.9$\pm$1.31} & \color{red}{51.81$\pm$3.74} & \color{red}{67.23$\pm$4.1} \\
    \midrule
    GRACE random & 83.49$\pm$2.36 & 89.34$\pm$2.39 & 64.96$\pm$2.32 & 56.56$\pm$2.67 & 46.5$\pm$2.82 & 62.59$\pm$2.56 \\
    GRACE optim & \color{violet}{86.2$\pm$2.32} & \color{violet}{90.84$\pm$1.92} & 70.4$\pm$1.93 & 57.17$\pm$1.99 & \color{violet}{47.86$\pm$2.26} & 62.53$\pm$1.99 \\
    \midrule
    CSGCL random & 83.91$\pm$2.18 & 88.77$\pm$1.51 & 73.41$\pm$2.2 & 44.24$\pm$2.41 & 42.12$\pm$1.73 & 55.82$\pm$4.02 \\
    CSGCL optim & 85.67$\pm$2.08 & \color{red}{\underline{91.54$\pm$1.63}} & \color{violet}{\underline{73.65$\pm$2.58}} & \color{violet}{\underline{60.58$\pm$2.09}} & \color{blue}{\underline{48.22$\pm$2.14}} & \color{blue}{\underline{65.4$\pm$2.95}} \\
    \midrule
    BGRL random & \color{blue}{86.81$\pm$1.54} & 89.27$\pm$1.0 & 60.46$\pm$3.09 & 52.03$\pm$2.92 & 40.24$\pm$2.77 & 48.83$\pm$2.55 \\
    BGRL optim & \color{red}{\underline{87.55$\pm$1.19}} & 88.64$\pm$1.6 & 69.55$\pm$2.59 & 54.85$\pm$2.12 & 43.6$\pm$2.4 & 59.57$\pm$3.42 \\
    \midrule
    L-GRACE random & 83.38$\pm$1.37 & 90.66$\pm$1.19 & OOM & OOM & OOM & OOM \\
    L-GRACE optim & 86.0$\pm$2.32 & \color{blue}{91.05$\pm$1.63} & OOM & OOM & OOM & OOM \\
    \midrule
    L-BGRL random & 83.89$\pm$2.56 & 90.62$\pm$1.35 & 63.44$\pm$3.13 & 46.42$\pm$1.71 & 38.1$\pm$3.03 & 56.5$\pm$3.14 \\
    L-BGRL optim & 83.68$\pm$2.11 & 90.15$\pm$1.75 & 62.12$\pm$3.26 & 46.78$\pm$2.04 & 41.16$\pm$2.91 & 61.98$\pm$3.47 \\
    \bottomrule
    \end{tabular}
    }
    \label{tab:attr-simple}
\end{table*}

We can observe that the L-GRACE method cannot run on cs, physics, computers, and photo due to an Out of Memory (greater than 96GB). In fact, the adaptation of GRACE uses the link representation, and then, the memory size does not depend anymore on the number of nodes, but on the number of links, which is greater than the number of nodes in most cases. Because of this scalable problem, we have to look at the L-BGRL to see the impact of the link representation contribution on the attributed graphs. Even if the adaptation of BGRL suffers from the same size complexity adaptation (depending on the number of links), BGRL was built for large-scale graphs, and then L-BGRL can scale up. However, as previously observed on non-attributed graphs, L-BGRL shows on-par performance with BGRL, but its performance deteriorates on attributed graphs. The use of link representation for attributed graphs doesn't seem appropriate, as the goal of link representation is to create invariant learning on the structure of links in the graph instead of nodes. 
This phenomenon has been discussed in Mao et al.'s work \cite{mao2023revisiting}, which shows that structural proximity and feature proximity can be potentially conflicting signals: \textit{"while structural proximity patterns may imply a likely link between two nodes, feature proximity patterns might suggest the opposite. Therefore, it seems challenging for a single model to benefit both node pairs with the feature proximity factor and those with the structural ones."}. The goal of our link loss prediction is to exploit the structural pattern on the graph to predict the link. Thus, the influence of attributes appears to be more negative than positive on the final performances of L-BGRL.

Comparing models' performances on attributed graphs can be inappropriate, as the final performances depend strongly on the attributes of nodes; for example, the difference in GCN's performance against other self-supervised ones, on cs and physics, can be explained by the phenomenon, while they have a large number of dimensions for attributes.

\section{Conclusion and Perspectives}

In this article, we explored the problem of self-supervised frameworks for link prediction, centered on two contributions in instance discrimination: treating the augmentation process as a tunable model parameter rather than a fixed choice, and developing link-based models (L-GRACE and L-BGRL) that directly learn from link representations rather than node embeddings. By adapting the loss function to compare link representations, L-GRACE and L-BGRL achieve state-of-the-art results on multiple datasets, demonstrating the effectiveness of structure-aware, task-aligned self-supervised learning for link prediction. The adaptation of the loss function to focus on link representations proves effective, particularly for non-attributed graphs, where structural information is paramount. Our extensive evaluation reveals that performance in link prediction is highly sensitive to the augmentation process. Our SBM-based augmentation, while concurrent in certain contexts, is constrained by the quality of community detection. 
The performance of self-supervised models for link prediction is highly sensitive to the choice of augmentation, and no single augmentation strategy is universally optimal across all datasets and models. 

A major limitation of our link loss is the scalability issue, which stems from the need to sample negative edges for contrastive learning. We adapt our link loss on BGRL to permit an evaluation on a larger graph, but it doesn't perform as well as L-GRACE. Many potential solutions exist, like adapting our link loss onto another self-supervised model, such as the Graph Group Discrimination (GGD) method \cite{zheng2022rethinking}, which is reported to be faster and less memory-consuming than BGRL. Another potential way to reduce the memory overhead is to sample a fixed-size set of neighbors (in the GraphSAGE manner \cite{hamilton2017inductive}) for each node at the construction of the $edge\_pos$ set, and thus reducing the number of link representations in the GPU memory. These research directions have to be addressed in future works to scale to larger graphs.

Future work should not only improve community-detection for SBM-based augmentation and scalability for large attributed graphs, but also systematically evaluate a broader range of augmentation strategies (beyond SBM) and identify why augmentation effectiveness varies across datasets (e.g., dataset topology, attribute quality, class imbalance, noise levels, etc.).
Additionally, further investigation into the interplay between node attributes and structural information could yield even more robust models for link prediction. It would be interesting to combine our new Adapted Link Loss with other Instance Discrimination methods. More generally, this work shows preliminary results that encourage research direction in Instance Discrimination learning for link prediction.
\bibliography{sn-bibliography}

\appendix
\section{Appendix} \label{appendix}
\begin{table*}[b]
    \centering
    \caption{Full link prediction results with the AP metric, on non-attributed graphs. Supervised methods are reported (sup). The top three models are colored by {\color{red}First}, {\color{blue}Second}, {\color{violet}Third}. Scores in bold are the best augmentation per method. $\bigstar$ shows the most significant best results, and X shows the most significant worst results from our experimentation.}
    \resizebox{\textwidth}{!}{
    \begin{tabular}{lllllllll}
    \toprule
     & USAir & NS & PB & Yeast & Celegans & Power & Router & Ecoli \\
    \midrule
    GCN (sup) & \color{blue}{96.35$\pm$0.6} & 92.4$\pm$0.53 & 94.34$\pm$0.33 & \color{blue}{95.37$\pm$0.23} & 84.91$\pm$0.86 & 71.38$\pm$0.9 & 66.71$\pm$2.17 X & 96.21$\pm$0.22 \\
    \midrule
    GRACE random & \textbf{95.93$\pm$0.74} & 92.81$\pm$0.35 & \textbf{94.19$\pm$0.3} & \textbf{95.04$\pm$0.24} & 87.72$\pm$0.83 & 73.4$\pm$1.03 & 70.64$\pm$1.48 & 95.93$\pm$0.23 \\
    GRACE deg & 94.76$\pm$2.92 & 92.96$\pm$1.08 & 93.65$\pm$0.57 & 94.74$\pm$0.25 & 86.79$\pm$1.3 & 71.49$\pm$9.17 & 61.3$\pm$9.81 X & 95.82$\pm$1.36 \\
    GRACE evc & 83.57$\pm$1.31 X & 89.98$\pm$0.98 & 92.59$\pm$0.41 & 94.66$\pm$0.17 & 87.32$\pm$1.13 & 73.86$\pm$1.04 & 68.68$\pm$3.32 & \color{violet}{\textbf{96.54$\pm$0.11}} \\
    GRACE pr & 95.56$\pm$0.65 & \textbf{94.39$\pm$0.6} & 94.12$\pm$0.35 & 94.8$\pm$0.23 & \textbf{87.98$\pm$0.59} & \color{blue}{\textbf{77.14$\pm$0.77 $\bigstar$}} & 85.48$\pm$0.86 & 96.38$\pm$0.17 \\
    GRACE scom & 94.86$\pm$0.69 & 92.48$\pm$0.55 & 93.54$\pm$0.31 & 94.62$\pm$0.22 & 86.45$\pm$0.42 & 76.35$\pm$0.89 & 80.69$\pm$1.57 & 96.43$\pm$0.18 \\
    GRACE sbm & 94.38$\pm$0.66 & 90.21$\pm$0.96 & 93.38$\pm$0.37 & 94.01$\pm$0.3 X & 85.17$\pm$1.12 & 75.54$\pm$1.14 & \textbf{75.01$\pm$1.76} & 96.04$\pm$0.14 \\
    GRACE sbm2 & 94.1$\pm$1.08 & 93.2$\pm$0.65 & 93.55$\pm$0.38 & 94.48$\pm$0.29 & 84.59$\pm$1.54 & 75.06$\pm$0.83 & 74.36$\pm$1.65 & 96.03$\pm$0.3 \\
    \midrule
    CSGCL random & 94.32$\pm$1.04 & 93.63$\pm$0.57 & 93.74$\pm$0.31 & 94.21$\pm$0.27 & 87.32$\pm$0.99 & 75.5$\pm$0.91 & 81.99$\pm$0.84 & 95.6$\pm$0.24 \\
    CSGCL deg & \textbf{95.38$\pm$1.01} & 92.7$\pm$1.13 & \textbf{94.15$\pm$0.27} & 94.64$\pm$0.24 & \textbf{88.06$\pm$1.35} & 73.47$\pm$2.34 & 60.0$\pm$10.16 X & 96.1$\pm$0.2 \\
    CSGCL pr & 93.8$\pm$0.96 & 92.43$\pm$0.84 & 91.6$\pm$0.48 X & 94.95$\pm$0.24 & 87.47$\pm$1.11 & 76.07$\pm$0.88 & 78.94$\pm$4.79 & 95.99$\pm$0.21 \\
    CSGCL evc & 92.71$\pm$1.13 & 89.92$\pm$1.2 & 94.06$\pm$0.35 & \textbf{94.98$\pm$0.27} & 87.57$\pm$0.72 & \textbf{76.2$\pm$0.74} & 67.68$\pm$2.18 & \textbf{96.1$\pm$0.3} \\
    CSGCL scom & 95.35$\pm$0.74 & \textbf{94.58$\pm$0.77} & 93.77$\pm$0.41 & 94.68$\pm$0.19 & 87.14$\pm$1.07 & 73.88$\pm$0.61 & \textbf{83.62$\pm$1.04} & 94.53$\pm$0.29 X \\
    CSGCL sbm & 94.39$\pm$0.97 & 92.07$\pm$0.49 & 92.88$\pm$0.44 & 94.21$\pm$0.29 & 80.4$\pm$1.77 X & 75.67$\pm$1.1 & 69.13$\pm$1.36 & 94.49$\pm$0.31 X \\
    CSGCL sbm2 & 95.08$\pm$0.74 & 91.49$\pm$0.71 & 93.47$\pm$0.28 & 93.81$\pm$0.33 X & 81.6$\pm$1.07 & 74.03$\pm$0.84 & 74.55$\pm$1.5 & 95.91$\pm$0.23 \\
    \midrule
    BGRL random & 94.62$\pm$0.78 & 92.59$\pm$0.63 & 93.62$\pm$0.37 & 94.59$\pm$0.22 & 86.02$\pm$0.98 & 67.98$\pm$0.69 & 69.21$\pm$1.32 & 96.08$\pm$0.21 \\
    BGRL deg & 94.07$\pm$1.14 & \color{blue}{\textbf{94.97$\pm$1.3}} & 93.54$\pm$0.61 & 94.49$\pm$0.28 & 87.13$\pm$1.59 & 73.76$\pm$2.96 & 77.1$\pm$4.96 & \color{blue}{\textbf{96.55$\pm$0.14}} \\
    BGRL evc & 94.94$\pm$0.63 & 92.46$\pm$0.53 & \textbf{94.02$\pm$0.35} & \textbf{94.76$\pm$0.24} & 84.67$\pm$1.52 & 67.9$\pm$0.57 & 85.09$\pm$2.91 & 96.09$\pm$0.19 \\
    BGRL pr & 94.4$\pm$0.77 & 92.02$\pm$0.48 & 93.28$\pm$0.39 & 94.56$\pm$0.31 & 86.13$\pm$1.3 & 70.25$\pm$1.3 & 85.49$\pm$0.5 & 95.91$\pm$0.22 \\
    BGRL scom & \textbf{95.26$\pm$0.67} & 93.51$\pm$0.37 & 93.34$\pm$0.35 & 94.66$\pm$0.23 & \textbf{87.15$\pm$0.76} & 69.5$\pm$0.78 & \color{red}{\textbf{88.57$\pm$0.68 $\bigstar$}} & 95.86$\pm$0.18 \\
    BGRL sbm & 94.98$\pm$0.8 & 92.29$\pm$0.69 & 92.63$\pm$0.34 & 93.98$\pm$0.39 X & 83.68$\pm$1.08 & \textbf{75.81$\pm$1.14} & \color{violet}{86.24$\pm$0.67} & 95.93$\pm$0.16 \\
    BGRL sbm2 & 92.31$\pm$1.35 & 93.96$\pm$0.49 & 93.19$\pm$0.41 & 94.12$\pm$0.32 & 83.39$\pm$1.29 & 72.96$\pm$1.32 & 81.18$\pm$1.11 & 95.92$\pm$0.28 \\
    \midrule
    L-GRACE random & \color{red}{\textbf{96.71$\pm$0.37 $\bigstar$}} & 92.18$\pm$0.56 & \color{red}{\textbf{94.74$\pm$0.3 $\bigstar$}} & \color{red}{\textbf{95.91$\pm$0.41 $\bigstar$}} & \color{violet}{88.35$\pm$1.02} & 72.86$\pm$0.94 & \textbf{72.83$\pm$8.82} & 96.09$\pm$0.24 \\
    L-GRACE deg & 95.7$\pm$0.9 & 92.82$\pm$0.68 & 93.49$\pm$0.44 & 94.75$\pm$0.31 & 86.47$\pm$1.27 & \color{red}{\textbf{78.24$\pm$4.13}} & 65.55$\pm$8.37 & \color{red}{\textbf{96.78$\pm$0.15 $\bigstar$}} \\
    L-GRACE evc & \color{violet}{96.27$\pm$0.72} & 93.03$\pm$0.45 & \color{violet}{94.39$\pm$0.34} & 94.82$\pm$0.23 & \color{red}{\textbf{88.98$\pm$1.13 $\bigstar$}} & 70.71$\pm$0.6 & 67.59$\pm$5.72 & 94.64$\pm$0.26 X \\
    L-GRACE pr & 95.47$\pm$0.82 & \color{red}{\textbf{95.58$\pm$0.44 $\bigstar$}} & 94.3$\pm$0.27 & \color{violet}{95.19$\pm$0.24} & 88.06$\pm$1.1 & 71.04$\pm$0.72 & 72.64$\pm$1.42 & 96.16$\pm$0.21 \\
    L-GRACE scom & 95.78$\pm$0.77 & 93.05$\pm$0.67 & \color{blue}{94.54$\pm$0.36 $\bigstar$} & 94.24$\pm$0.87 & 85.5$\pm$1.15 & \color{violet}{76.5$\pm$0.96} & 68.86$\pm$1.12 & 96.47$\pm$0.18 \\
    L-GRACE sbm & 95.36$\pm$0.87 & 77.92$\pm$11.97 X & 93.09$\pm$0.3 & 94.94$\pm$0.3 & 86.64$\pm$0.92 & 73.45$\pm$0.58 & 72.62$\pm$2.76 & 96.44$\pm$0.18 \\
    L-GRACE sbm2 & 94.82$\pm$0.91 & 91.32$\pm$1.3 & 93.13$\pm$0.68 & 94.52$\pm$0.35 & 84.3$\pm$1.77 & 73.99$\pm$2.03 & 66.44$\pm$2.09 X & 96.4$\pm$0.26 \\
    \midrule
    L-BGRL random & 91.36$\pm$1.2 & 92.54$\pm$0.57 & 93.98$\pm$0.29 & 95.12$\pm$0.34 & 85.76$\pm$1.29 & 63.55$\pm$1.16 X & 79.67$\pm$5.96 & 95.93$\pm$0.2 \\
    L-BGRL deg & 93.14$\pm$1.36 & 94.7$\pm$1.06 & 93.23$\pm$0.38 & 94.83$\pm$0.15 & \color{blue}{\textbf{88.72$\pm$0.77 $\bigstar$}} & \textbf{73.66$\pm$3.97} & \color{blue}{\textbf{86.46$\pm$0.67}} & 95.29$\pm$0.19 \\
    L-BGRL evc & 95.49$\pm$0.72 & 88.25$\pm$1.5 X & 91.03$\pm$0.45 X & 94.97$\pm$0.23 & 86.82$\pm$0.98 & 66.1$\pm$1.4 & 74.74$\pm$4.38 & \textbf{96.35$\pm$0.1} \\
    L-BGRL pr & \textbf{95.95$\pm$0.97} & \color{violet}{\textbf{94.97$\pm$0.7 $\bigstar$}} & \textbf{94.03$\pm$0.24} & \textbf{95.14$\pm$0.19} & 87.79$\pm$1.03 & 67.74$\pm$1.91 & 83.06$\pm$3.35 & 96.06$\pm$0.26 \\
    L-BGRL scom & 94.86$\pm$0.82 & 91.55$\pm$0.88 & 93.77$\pm$0.35 & 94.39$\pm$0.39 & 87.49$\pm$0.86 & 63.21$\pm$0.99 X & 72.79$\pm$1.54 & 96.25$\pm$0.26 \\
    L-BGRL sbm & 91.9$\pm$1.11 & 94.95$\pm$0.73 & 92.61$\pm$0.32 & 94.55$\pm$0.3 & 85.91$\pm$0.72 & 73.45$\pm$0.93 & 80.89$\pm$1.76 & 95.56$\pm$0.17 \\
    L-BGRL sbm2 & 94.62$\pm$0.94 & 93.06$\pm$0.76 & 93.54$\pm$0.22 & 94.05$\pm$0.3 & 74.65$\pm$1.21 X & 70.1$\pm$1.72 & 77.46$\pm$1.28 & 96.02$\pm$0.23 \\
    \bottomrule
    \end{tabular}
    
    }
    \label{tab:non-attr AP}
\end{table*}
\begin{table*}[]
    \centering
    \caption{Full link prediction results with the AUC metric, on non-attributed graphs. Supervised methods are reported (sup). The top three models are colored by {\color{red}First}, {\color{blue}Second}, {\color{violet}Third}. Scores in bold are the best augmentation per method. $\bigstar$ shows the most significant best results, and X shows the most significant worst results from our experimentation.}
    \resizebox{\textwidth}{!}{
    \begin{tabular}{lllllllll}
    \toprule
     & USAir & NS & PB & Yeast & Celegans & Power & Router & Ecoli \\
    \midrule
    GCN (sup)& \color{blue}{95.8$\pm$0.69 $\bigstar$} & 87.68$\pm$1.03 & 94.06$\pm$0.26 & \color{blue}{93.01$\pm$0.39} & 84.93$\pm$1.11 & 64.84$\pm$1.24 & 58.77$\pm$3.39 & 94.68$\pm$0.29 \\
    \midrule
    GRACE random & \textbf{94.23$\pm$1.12} & 89.19$\pm$0.58 & \textbf{94.19$\pm$0.24} & \textbf{92.35$\pm$0.27} & 88.17$\pm$0.97 & 66.52$\pm$1.0 & 62.77$\pm$1.82 & 94.46$\pm$0.25 \\
    GRACE deg & 93.26$\pm$3.04 & 89.08$\pm$1.8 & 93.44$\pm$0.4 & 91.58$\pm$0.45 & 87.21$\pm$1.37 & 67.02$\pm$9.46 & 53.23$\pm$11.34 X & 94.57$\pm$1.06 \\
    GRACE evc & 82.86$\pm$1.34 X & 87.42$\pm$1.28 & 92.11$\pm$0.19 & 91.51$\pm$0.27 & 88.85$\pm$0.86 & 68.82$\pm$1.09 & 64.09$\pm$4.1 & \color{blue}{\textbf{95.42$\pm$0.11 $\bigstar$}} \\
    GRACE pr & 93.94$\pm$1.08 & 92.05$\pm$0.91 & 93.97$\pm$0.22 & 92.1$\pm$0.35 & \color{blue}{\textbf{89.61$\pm$0.66 $\bigstar$}} & \color{violet}{\textbf{73.46$\pm$0.93 $\bigstar$}} & \textbf{80.68$\pm$1.23} & 95.18$\pm$0.29 \\
    GRACE scom & 93.84$\pm$1.23 & 88.08$\pm$0.92 & 93.53$\pm$0.3 & 91.53$\pm$0.29 & 86.69$\pm$0.5 & 70.52$\pm$1.31 & 75.37$\pm$2.46 & 95.22$\pm$0.23 \\
    GRACE sbm & 92.7$\pm$1.02 & 87.6$\pm$1.04 & 93.09$\pm$0.25 & 90.76$\pm$0.29 & 85.29$\pm$1.34 & 72.5$\pm$1.53 & 68.08$\pm$2.69 & 94.49$\pm$0.3 \\
    GRACE sbm2 & 92.23$\pm$1.48 & \textbf{90.73$\pm$0.81} & 93.41$\pm$0.34 & 91.57$\pm$0.37 & 86.84$\pm$1.22 & 69.87$\pm$0.67 & 71.22$\pm$2.01 & 94.8$\pm$0.28 \\
    \midrule
    CSGCL random & 93.07$\pm$1.32 & 90.4$\pm$1.04 & 93.39$\pm$0.19 & 90.96$\pm$0.33 & 88.68$\pm$0.93 & 69.84$\pm$1.27 & 79.05$\pm$0.75 & 94.02$\pm$0.21 \\
    CSGCL deg & \textbf{93.68$\pm$1.41} & 89.66$\pm$0.89 & \textbf{93.84$\pm$0.2} & 91.49$\pm$0.45 & 88.75$\pm$1.26 & 66.11$\pm$3.38 & 48.51$\pm$13.43 X & 94.39$\pm$0.35 \\
    CSGCL pr & 91.85$\pm$1.28 & 89.97$\pm$1.22 & 91.09$\pm$0.34 X & \textbf{92.52$\pm$0.24} & \textbf{89.15$\pm$1.08} & 71.38$\pm$1.08 & 76.52$\pm$6.32 & 94.79$\pm$0.22 \\
    CSGCL evc & 90.91$\pm$1.43 & 86.11$\pm$2.02 & 93.66$\pm$0.26 & 92.37$\pm$0.32 & 89.07$\pm$0.77 & \textbf{71.95$\pm$1.04} & 57.57$\pm$4.07 & \textbf{94.98$\pm$0.32} \\
    CSGCL scom & 93.45$\pm$1.26 & \textbf{92.38$\pm$1.21} & 93.62$\pm$0.31 & 91.63$\pm$0.28 & 87.44$\pm$1.03 & 66.92$\pm$0.54 & \textbf{80.04$\pm$1.19} & 92.98$\pm$0.36 \\
    CSGCL sbm & 92.29$\pm$1.38 & 87.4$\pm$0.65 & 92.66$\pm$0.26 & 91.2$\pm$0.42 & 82.22$\pm$1.39 & 70.8$\pm$1.31 & 60.58$\pm$2.21 & 92.72$\pm$0.32 X \\
    CSGCL sbm2 & 93.45$\pm$1.12 & 86.54$\pm$0.94 & 93.48$\pm$0.26 & 90.61$\pm$0.35 X & 81.54$\pm$0.97 X & 67.68$\pm$0.84 & 71.14$\pm$1.28 & 94.58$\pm$0.28 \\
    \midrule
    BGRL random & 92.94$\pm$1.14 & 88.35$\pm$1.13 & 93.24$\pm$0.28 & 91.23$\pm$0.35 & 86.6$\pm$1.09 & 60.49$\pm$0.89 & 58.62$\pm$2.0 & 94.44$\pm$0.22 \\
    BGRL deg & 92.24$\pm$1.51 & \color{violet}{\textbf{92.9$\pm$2.21}} & 93.11$\pm$0.59 & 91.29$\pm$0.47 & 87.61$\pm$1.46 & 69.23$\pm$3.85 & 70.12$\pm$7.12 & \textbf{95.1$\pm$0.19} \\
    BGRL evc & 92.71$\pm$1.16 & 88.03$\pm$1.01 & \textbf{93.66$\pm$0.3} & \textbf{91.66$\pm$0.33} & 85.92$\pm$1.48 & 63.06$\pm$0.75 & 81.88$\pm$4.67 & 94.57$\pm$0.26 \\
    BGRL pr & 92.49$\pm$1.14 & 87.14$\pm$0.87 & 93.07$\pm$0.28 & 91.34$\pm$0.42 & 87.89$\pm$1.08 & 63.96$\pm$1.98 & 82.55$\pm$0.41 & 94.5$\pm$0.23 \\
    BGRL scom & \textbf{93.32$\pm$1.01} & 90.22$\pm$0.77 & 93.18$\pm$0.28 & 91.53$\pm$0.29 & \textbf{88.15$\pm$0.88} & 63.55$\pm$0.94 & \color{red}{\textbf{87.48$\pm$0.72 $\bigstar$}} & 94.33$\pm$0.28 \\
    BGRL sbm & 93.09$\pm$1.16 & 89.1$\pm$0.98 & 92.06$\pm$0.37 & 90.7$\pm$0.43 X & 85.37$\pm$0.99 & \textbf{71.56$\pm$1.22} & \color{violet}{83.74$\pm$0.86} & 94.42$\pm$0.2 \\
    BGRL sbm2 & 89.8$\pm$1.86 & 91.84$\pm$0.71 & 93.26$\pm$0.22 & 90.91$\pm$0.38 & 85.58$\pm$0.93 & 67.7$\pm$1.49 & 75.56$\pm$1.3 & 94.59$\pm$0.36 \\
    \midrule
    L-GRACE random & \color{red}{\textbf{96.16$\pm$0.53 $\bigstar$}} & 87.25$\pm$0.91 & \color{red}{\textbf{94.65$\pm$0.14 $\bigstar$}} & \color{red}{\textbf{94.02$\pm$0.81 $\bigstar$}} & \color{violet}{\textbf{89.6$\pm$0.95 $\bigstar$}} & 66.89$\pm$1.14 & 65.6$\pm$12.11 & 94.52$\pm$0.3 \\
    L-GRACE deg & 93.96$\pm$1.58 & 89.26$\pm$1.24 & 93.02$\pm$0.74 & 91.83$\pm$0.39 & 87.08$\pm$1.07 & \color{red}{\textbf{74.43$\pm$6.27}} & 52.77$\pm$13.25 & \color{red}{\textbf{95.47$\pm$0.32 $\bigstar$}} \\
    L-GRACE evc & \color{violet}{95.58$\pm$1.0 $\bigstar$} & 89.85$\pm$0.78 & 94.14$\pm$0.28 & 91.92$\pm$0.34 & 89.56$\pm$1.36 & 63.76$\pm$0.85 & 58.51$\pm$7.34 & 91.93$\pm$0.46 X \\
    L-GRACE pr & 93.88$\pm$1.43 & \color{red}{\textbf{94.38$\pm$0.68 $\bigstar$}} & \color{violet}{94.29$\pm$0.14} & 92.53$\pm$0.4 & 88.96$\pm$1.18 & 64.63$\pm$0.95 & 63.99$\pm$1.81 & 94.52$\pm$0.39 \\
    L-GRACE scom & 93.91$\pm$1.31 & 89.39$\pm$1.19 & \color{blue}{94.51$\pm$0.25 $\bigstar$} & 92.48$\pm$0.59 & 86.19$\pm$1.23 & \color{blue}{73.98$\pm$1.04 $\bigstar$} & 58.06$\pm$1.61 X & 95.03$\pm$0.18 \\
    L-GRACE sbm & 93.55$\pm$1.25 & 72.09$\pm$15.8 X & 92.3$\pm$0.36 & 92.37$\pm$0.46 & 87.63$\pm$0.86 & 66.64$\pm$0.79 & \textbf{66.25$\pm$4.19} & 95.08$\pm$0.27 \\
    L-GRACE sbm2 & 92.93$\pm$1.35 & 88.65$\pm$1.14 & 92.8$\pm$0.32 & 91.65$\pm$0.51 & 85.27$\pm$1.46 & 66.56$\pm$3.16 & 54.66$\pm$3.76 X & \color{violet}{95.38$\pm$0.26 $\bigstar$} \\
    \midrule
    L-BGRL random & 88.55$\pm$1.49 X & 88.58$\pm$1.01 & 93.73$\pm$0.26 & \color{violet}{\textbf{92.74$\pm$0.61}} & 87.32$\pm$1.1 & 52.43$\pm$2.78 X & 74.77$\pm$8.09 & 94.03$\pm$0.32 \\
    L-BGRL deg & 90.82$\pm$1.68 & \color{blue}{\textbf{93.09$\pm$1.92}} & 92.93$\pm$0.28 & 91.99$\pm$0.25 & \color{red}{\textbf{89.8$\pm$0.91 $\bigstar$}} & \textbf{68.19$\pm$5.12} & \color{blue}{\textbf{84.26$\pm$0.86}} & 94.03$\pm$0.31 \\
    L-BGRL evc & \textbf{94.53$\pm$0.87} & 85.14$\pm$1.46 X & 91.71$\pm$0.2 & 91.92$\pm$0.37 & 88.47$\pm$0.99 & 55.25$\pm$3.19 & 67.51$\pm$6.79 & 94.78$\pm$0.21 \\
    L-BGRL pr & 94.52$\pm$1.44 & 92.66$\pm$1.13 & \textbf{93.78$\pm$0.18} & 92.42$\pm$0.29 & 89.19$\pm$0.86 & 63.16$\pm$2.9 & 79.01$\pm$5.34 & 94.36$\pm$0.24 \\
    L-BGRL scom & 93.64$\pm$1.24 & 87.55$\pm$1.57 & 93.61$\pm$0.24 & 91.16$\pm$0.47 & 88.4$\pm$0.93 & 56.2$\pm$2.04 X & 63.74$\pm$2.33 & \textbf{94.81$\pm$0.29} \\
    L-BGRL sbm & 90.3$\pm$1.59 & 92.81$\pm$1.22 & 92.11$\pm$0.23 & 91.96$\pm$0.43 & 87.67$\pm$0.83 & 65.8$\pm$1.43 & 75.68$\pm$2.51 & 93.97$\pm$0.27 \\
    L-BGRL sbm2 & 92.59$\pm$1.48 & 88.85$\pm$1.35 & 93.55$\pm$0.14 & 90.58$\pm$0.32 X & 74.98$\pm$1.11 X & 60.54$\pm$2.54 & 73.77$\pm$1.48 & 94.69$\pm$0.19 \\
    \bottomrule
    \end{tabular}
    
    }
    \label{tab:non-attr AUC}
\end{table*}
\begin{table*}[]
    \centering
    \caption{Link prediction results with Hits@50 metric, on attributed graphs. Reported methods from the T-BGRL article are noted with *. Supervised methods are reported (sup). The top three models are colored by {\color{red}First}, {\color{blue}Second}, {\color{violet}Third}. Scores in bold are the best augmentation per method. $\bigstar$ shows the most significant best results, and X shows the most significant worst results from our experimentation. The best scores from the SSL model are underlined.}
    \resizebox{\textwidth}{!}{
    \begin{tabular}{lllllll}
    \toprule
     & cora & citeseer & cs & physics & computers & photo \\
    \midrule
    GCN* (sup) & 81.6$\pm$1.3 & 82.2$\pm$1.17 & \color{blue}{76.2$\pm$1} & \color{blue}{79.8$\pm$1.8} & 42.6$\pm$3.6 & \color{violet}{64.2$\pm$2.9} \\
    GRACE random* & 68.6$\pm$5.6 & 70.7$\pm$6.8 & 45.6$\pm$6.6 & OOM & 24$\pm$2.7 & 48.6$\pm$2.5 \\
    BGRL random* & 79.2$\pm$1.5 & 85.8$\pm$2 & 51.5$\pm$1.6 & 47.6$\pm$1.5 & 34.6$\pm$1.8 & 56.2$\pm$1.3 \\
    T-BGRL random* & 77.3$\pm$2 & 86.8$\pm$2.3 & 55.5$\pm$0.9 & 47.1$\pm$2.1 & 31.5$\pm$1.5 & 51.7$\pm$1.6 \\
    \midrule
    GCN (sup) & 81.97$\pm$2.31 & 75.01$\pm$3.03 X & \color{red}{83.89$\pm$1.28 $\bigstar$} & \color{red}{83.9$\pm$1.31 $\bigstar$} & \color{red}{51.81$\pm$3.74 $\bigstar$} & \color{red}{67.23$\pm$4.1 $\bigstar$} \\
    \midrule
    GRACE random & 83.49$\pm$2.36 & 89.34$\pm$2.39 & 64.96$\pm$2.32 & 56.56$\pm$2.67 & 46.5$\pm$2.82 & \textbf{62.59$\pm$2.56} \\
    GRACE deg & 81.94$\pm$3.06 & 86.62$\pm$2.19 & \textbf{70.4$\pm$1.93} & 39.64$\pm$2.49 & \color{violet}{\textbf{47.86$\pm$2.26 $\bigstar$}} & 62.53$\pm$1.99 \\
    GRACE evc & 85.28$\pm$1.75 & 84.95$\pm$2.85 & 68.23$\pm$2.38 & 41.46$\pm$1.97 & 40.9$\pm$3.5 & 59.97$\pm$2.49 \\
    GRACE pr & 80.15$\pm$2.15 & 88.44$\pm$1.02 & 62.58$\pm$2.08 & 48.0$\pm$3.04 & 45.53$\pm$2.76 & 33.95$\pm$4.42 X \\
    GRACE scom & \color{violet}{\textbf{86.2$\pm$2.32}} & 90.64$\pm$1.58 $\bigstar$ & 70.38$\pm$2.74 & \textbf{57.17$\pm$1.99} & 36.87$\pm$1.95 & 56.45$\pm$3.13 \\
    GRACE sbm & 78.22$\pm$3.35 & 88.46$\pm$1.71 & 60.75$\pm$2.82 & 43.76$\pm$2.35 & 36.16$\pm$2.61 & 56.31$\pm$3.84 \\
    GRACE sbm2 & 83.8$\pm$1.8 & \color{violet}{\textbf{90.84$\pm$1.92 $\bigstar$}} & 36.94$\pm$1.28 X & 54.94$\pm$3.27 & 39.55$\pm$2.63 & 49.37$\pm$3.18 \\
    \midrule
    CSGCL random & 83.91$\pm$2.18 & 88.77$\pm$1.51 & 73.41$\pm$2.2 $\bigstar$ & 44.24$\pm$2.41 & 42.12$\pm$1.73 & 55.82$\pm$4.02 \\
    CSGCL deg & 85.37$\pm$1.17 & 87.87$\pm$1.72 & \color{violet}{\textbf{\underline{73.65$\pm$2.58}}} & 53.22$\pm$2.54 & 40.23$\pm$2.33 & 60.37$\pm$2.47 \\
    CSGCL evc & 83.66$\pm$2.63 & 84.04$\pm$1.17 & 61.87$\pm$2.83 & 54.6$\pm$2.54 & \color{blue}{\textbf{\underline{48.22$\pm$2.14 $\bigstar$}}} & 55.07$\pm$2.19 \\
    CSGCL pr & 85.43$\pm$2.14 & \color{red}{\textbf{\underline{91.54$\pm$1.63 $\bigstar$}}} & 64.04$\pm$2.36 & 51.23$\pm$2.29 & 38.71$\pm$2.03 & 62.72$\pm$3.16 \\
    CSGCL scom & 81.4$\pm$2.81 & 86.9$\pm$1.41 & 66.21$\pm$2.21 & \color{violet}{\textbf{\underline{60.58$\pm$2.09 $\bigstar$}}} & 31.39$\pm$2.01 X & \color{blue}{\textbf{\underline{65.4$\pm$2.95 $\bigstar$}}} \\
    CSGCL sbm & \textbf{85.67$\pm$2.08} & 82.77$\pm$2.08 & 58.36$\pm$2.88 & 46.56$\pm$2.44 & 37.57$\pm$2.25 & 56.28$\pm$2.98 \\
    CSGCL sbm2 & 80.15$\pm$3.18 & 88.44$\pm$1.72 & 59.71$\pm$1.71 & 41.51$\pm$2.0 & 34.39$\pm$3.14 & 48.97$\pm$2.4 \\
    \midrule
    BGRL random & \color{blue}{86.81$\pm$1.54 $\bigstar$} & \textbf{89.27$\pm$1.0} & 60.46$\pm$3.09 & 52.03$\pm$2.92 & 40.24$\pm$2.77 & 48.83$\pm$2.55 \\
    BGRL deg & 85.67$\pm$2.09 & 82.48$\pm$2.12 & \textbf{69.55$\pm$2.59} & 38.94$\pm$2.06 & 41.8$\pm$2.55 & 59.18$\pm$2.81 \\
    BGRL evc & \color{red}{\textbf{\underline{87.55$\pm$1.19 $\bigstar$}}} & 83.23$\pm$3.03 & 69.17$\pm$2.77 & 47.39$\pm$2.43 & 40.48$\pm$2.34 & \textbf{59.57$\pm$3.42} \\
    BGRL pr & 76.87$\pm$3.55 X & 85.78$\pm$2.65 & 67.8$\pm$2.4 & 38.59$\pm$3.02 & \textbf{43.6$\pm$2.4} & 57.43$\pm$3.14 \\
    BGRL scom & 83.11$\pm$1.49 & 82.99$\pm$1.94 & 66.69$\pm$3.06 & \textbf{54.85$\pm$2.12} & 41.51$\pm$2.03 & 52.05$\pm$4.06 \\
    BGRL sbm & 83.83$\pm$1.77 & 88.64$\pm$1.6 & 60.18$\pm$2.4 & 41.91$\pm$3.11 & 36.44$\pm$2.2 & 54.34$\pm$4.41 \\
    BGRL sbm2 & 81.35$\pm$2.35 & 81.67$\pm$2.64 & 55.05$\pm$2.61 & 46.64$\pm$4.49 & 38.33$\pm$2.71 & 48.25$\pm$3.79 \\
    \midrule
    L-GRACE random & 83.38$\pm$1.37 & 90.66$\pm$1.19 & OOM & OOM & OOM & OOM \\
    L-GRACE deg & 82.33$\pm$2.98 & 86.11$\pm$1.87 & OOM & OOM & OOM & OOM \\
    L-GRACE evc & 76.64$\pm$3.3 X & 88.24$\pm$1.46 & OOM & OOM & OOM & OOM \\
    L-GRACE pr & 84.76$\pm$1.98 & \color{blue}{\textbf{91.05$\pm$1.63 $\bigstar$}} & OOM & OOM & OOM & OOM \\
    L-GRACE scom & \textbf{86.0$\pm$2.32} & 79.45$\pm$2.87 & OOM & OOM & OOM & OOM \\
    L-GRACE sbm & 83.72$\pm$2.07 & 80.29$\pm$2.13 & OOM & OOM & OOM & OOM \\
    L-GRACE sbm2 & 80.3$\pm$2.89 & 76.59$\pm$4.42 X & OOM & OOM & OOM & OOM \\
    \midrule
    L-BGRL random & \textbf{83.89$\pm$2.56} & \textbf{90.62$\pm$1.35} & \textbf{63.44$\pm$3.13} & 46.42$\pm$1.71 & 38.1$\pm$3.03 & 56.5$\pm$3.14 \\
    L-BGRL deg & 77.7$\pm$2.88 X & 86.81$\pm$2.05 & 62.12$\pm$3.26 & \textbf{46.78$\pm$2.04} & \textbf{41.16$\pm$2.91} & 52.99$\pm$3.19 \\
    L-BGRL evc & 83.68$\pm$2.11 & 85.3$\pm$2.34 & 56.85$\pm$2.42 & 41.53$\pm$1.76 & 38.05$\pm$3.09 & 61.86$\pm$3.57 \\
    L-BGRL pr & 83.02$\pm$2.33 & 88.33$\pm$1.61 & 61.7$\pm$2.08 & 38.63$\pm$3.21 & 39.75$\pm$2.64 & \textbf{61.98$\pm$3.47} \\
    L-BGRL scom & 79.72$\pm$3.15 & 90.15$\pm$1.75 & 56.26$\pm$3.61 & 41.22$\pm$1.51 & 32.58$\pm$4.06 X & 51.26$\pm$2.92 \\
    L-BGRL sbm & 82.81$\pm$2.0 & 87.19$\pm$2.18 & 55.41$\pm$2.87 & 32.69$\pm$2.15 X & 38.52$\pm$2.51 & 47.69$\pm$4.21 \\
    L-BGRL sbm2 & 83.23$\pm$2.0 & 88.79$\pm$1.9 & 54.77$\pm$1.68 & 39.5$\pm$0.7 & 37.98$\pm$3.04 & 36.29$\pm$4.56 X \\
    \bottomrule
    \end{tabular}
    }
    \label{tab:attr}
\end{table*}
\begin{table*}[]
    \centering
    \caption{Link prediction results with AP metric, on attributed graphs. Reported methods from the T-BGRL article are noted with *. Supervised methods are reported (sup). The top three models are colored by {\color{red}First}, {\color{blue}Second}, {\color{violet}Third}. Scores in bold are the best augmentation per method. $\bigstar$ shows the most significant best results, and X shows the most significant worst results from our experimentation.}
    \resizebox{\textwidth}{!}{
    \begin{tabular}{lllllll}
    \toprule
     & cora & citeseer & cs & physics & computers & photo \\
    \midrule
    GCN (sup) & 93.89$\pm$0.8 & 91.68$\pm$0.93 X & \color{red}{98.28$\pm$0.08 $\bigstar$} & \color{red}{99.16$\pm$0.02 $\bigstar$} & \color{red}{98.85$\pm$0.07 $\bigstar$} & \color{red}{99.05$\pm$0.07 $\bigstar$} \\
    \midrule
    GRACE random & 94.23$\pm$0.65 & 95.75$\pm$0.83 & 97.93$\pm$0.08 & 98.33$\pm$0.05 & 98.69$\pm$0.05 & 98.84$\pm$0.1 \\
    GRACE deg & 93.27$\pm$0.63 & 94.8$\pm$0.73 & \textbf{97.98$\pm$0.1} & 97.85$\pm$0.05 & \color{violet}{\textbf{98.81$\pm$0.05 $\bigstar$}} & \color{violet}{\textbf{98.92$\pm$0.06}} \\
    GRACE evc & 94.84$\pm$0.72 & 93.98$\pm$0.86 & 97.81$\pm$0.11 & 97.55$\pm$0.09 & 98.58$\pm$0.06 & 98.79$\pm$0.08 \\
    GRACE pr & 93.21$\pm$0.82 & 95.75$\pm$0.51 & 97.83$\pm$0.1 & 98.08$\pm$0.07 & 98.72$\pm$0.05 & 96.51$\pm$0.24 X \\
    GRACE scom & \textbf{94.86$\pm$0.58} & \textbf{96.52$\pm$0.62} & 97.89$\pm$0.11 & \color{violet}{\textbf{98.58$\pm$0.05}} & 98.21$\pm$0.1 & 98.67$\pm$0.08 \\
    GRACE sbm & 91.85$\pm$1.5 & 95.67$\pm$0.79 & 97.71$\pm$0.16 & 98.16$\pm$0.08 & 98.09$\pm$0.07 & 98.63$\pm$0.11 \\
    GRACE sbm2 & 94.24$\pm$0.5 & 96.42$\pm$0.61 & 95.6$\pm$0.15 X & 98.54$\pm$0.04 & 98.37$\pm$0.06 & 98.29$\pm$0.11 \\
    \midrule
    CSGCL random & 94.01$\pm$0.71 & 95.15$\pm$1.0 & 97.49$\pm$0.12 & 98.13$\pm$0.05 & 98.66$\pm$0.05 & 98.65$\pm$0.12 \\
    CSGCL deg & 94.64$\pm$0.67 & 95.82$\pm$0.61 & 97.84$\pm$0.09 & 98.56$\pm$0.05 & 98.48$\pm$0.06 & 98.79$\pm$0.08 \\
    CSGCL evc & 94.23$\pm$0.89 & 94.06$\pm$0.79 & 97.79$\pm$0.12 & 98.43$\pm$0.05 & \color{blue}{\textbf{98.83$\pm$0.06 $\bigstar$}} & 98.55$\pm$0.07 \\
    CSGCL pr & 94.84$\pm$0.82 & \color{blue}{\textbf{96.64$\pm$0.58 $\bigstar$}} & 97.72$\pm$0.1 & \color{blue}{\textbf{98.67$\pm$0.05 $\bigstar$}} & 98.37$\pm$0.06 & \textbf{98.92$\pm$0.07} \\
    CSGCL scom & 93.41$\pm$0.78 & 95.53$\pm$0.58 & \color{blue}{\textbf{98.05$\pm$0.11}} & 98.37$\pm$0.06 & 97.49$\pm$0.1 X & 98.88$\pm$0.06 \\
    CSGCL sbm & \color{violet}{\textbf{94.88$\pm$0.53}} & 93.73$\pm$0.88 & 97.22$\pm$0.11 & 98.31$\pm$0.08 & 98.35$\pm$0.07 & 98.6$\pm$0.08 \\
    CSGCL sbm2 & 93.1$\pm$0.88 & 95.69$\pm$0.74 & 96.87$\pm$0.12 X & 97.26$\pm$0.1 X & 98.15$\pm$0.08 & 98.32$\pm$0.08 \\
    \midrule
    BGRL random & \color{blue}{95.19$\pm$0.65} & 96.16$\pm$0.5 & 97.76$\pm$0.12 & 98.53$\pm$0.06 & 98.5$\pm$0.07 & 98.26$\pm$0.1 \\
    BGRL deg & 94.86$\pm$0.55 & 93.51$\pm$1.05 & 97.75$\pm$0.14 & 98.25$\pm$0.05 & 98.52$\pm$0.05 & \textbf{98.78$\pm$0.11} \\
    BGRL evc & \color{red}{\textbf{95.61$\pm$0.5 $\bigstar$}} & 93.71$\pm$0.9 & \color{violet}{\textbf{98.04$\pm$0.1}} & 97.96$\pm$0.1 & 98.53$\pm$0.06 & 98.74$\pm$0.09 \\
    BGRL pr & 91.7$\pm$0.91 X & 94.98$\pm$0.81 & 97.41$\pm$0.1 & 97.88$\pm$0.08 & \textbf{98.58$\pm$0.08} & 98.67$\pm$0.09 \\
    BGRL scom & 93.99$\pm$0.72 & 93.76$\pm$0.83 & 97.85$\pm$0.12 & \textbf{98.54$\pm$0.04} & 98.56$\pm$0.06 & 98.37$\pm$0.11 \\
    BGRL sbm & 94.36$\pm$0.34 & \textbf{95.49$\pm$0.69} & 97.43$\pm$0.11 & 97.84$\pm$0.06 & 98.26$\pm$0.06 & 98.52$\pm$0.13 \\
    BGRL sbm2 & 93.46$\pm$0.69 & 93.15$\pm$0.84 & 96.89$\pm$0.11 & 97.96$\pm$0.11 & 98.3$\pm$0.04 & 98.16$\pm$0.14 \\
    \midrule
    L-GRACE random & 93.2$\pm$0.82 & 96.34$\pm$0.54 & OOM & OOM & OOM & OOM \\
    L-GRACE deg & 94.15$\pm$0.98 & 94.69$\pm$0.76 & OOM & OOM & OOM & OOM \\
    L-GRACE evc & 91.87$\pm$0.97 X & 95.79$\pm$0.56 & OOM & OOM & OOM & OOM \\
    L-GRACE pr & 94.56$\pm$0.71 & \color{red}{\textbf{96.74$\pm$0.59 $\bigstar$}} & OOM & OOM & OOM & OOM \\
    L-GRACE scom & \textbf{94.84$\pm$0.52} & 92.89$\pm$1.12 & OOM & OOM & OOM & OOM \\
    L-GRACE sbm & 94.0$\pm$0.73 & 93.32$\pm$0.61 & OOM & OOM & OOM & OOM \\
    L-GRACE sbm2 & 92.95$\pm$0.65 & 91.83$\pm$1.25 X & OOM & OOM & OOM & OOM \\
    \midrule
    L-BGRL random & \textbf{94.38$\pm$1.01} & \color{violet}{\textbf{96.62$\pm$0.51}} & \textbf{97.79$\pm$0.09} & \textbf{98.42$\pm$0.05} & 98.45$\pm$0.06 & 98.74$\pm$0.14 \\
    L-BGRL deg & 92.45$\pm$0.93 & 95.6$\pm$0.7 & 97.74$\pm$0.12 & 97.98$\pm$0.05 & 98.43$\pm$0.06 & 98.56$\pm$0.12 \\
    L-BGRL evc & 94.22$\pm$0.53 & 94.92$\pm$0.87 & 97.59$\pm$0.12 & 98.07$\pm$0.08 & 98.26$\pm$0.1 & \color{blue}{\textbf{98.94$\pm$0.1}} \\
    L-BGRL pr & 94.19$\pm$0.52 & 95.92$\pm$0.73 & 97.44$\pm$0.08 & 97.96$\pm$0.07 & \textbf{98.49$\pm$0.06} & 98.86$\pm$0.09 \\
    L-BGRL scom & 93.08$\pm$0.84 & 96.38$\pm$0.76 & 97.52$\pm$0.16 & 97.6$\pm$0.09 & 98.01$\pm$0.12 X & 98.3$\pm$0.1 \\
    L-BGRL sbm & 94.2$\pm$0.73 & 95.82$\pm$0.59 & 97.29$\pm$0.17 & 97.26$\pm$0.07 X & 98.37$\pm$0.08 & 98.04$\pm$0.25 \\
    L-BGRL sbm2 & 94.02$\pm$0.64 & 95.91$\pm$0.7 & 97.31$\pm$0.11 & 97.61$\pm$0.09 & 98.33$\pm$0.06 & 96.84$\pm$0.47 X \\
    \bottomrule
    \end{tabular}
    }
    \label{tab:attr AP}
\end{table*}
\begin{table*}[]
    \centering
    \caption{Link prediction results with ROC-AUC metric, on attributed graphs. Reported methods from the T-BGRL article are noted with *. Supervised methods are reported (sup). The top three models are colored by {\color{red}First}, {\color{blue}Second}, {\color{violet}Third}. Scores in bold are the best augmentation per method. $\bigstar$ shows the most significant best results, and X shows the most significant worst results from our experimentation.}
    \resizebox{\textwidth}{!}{
    \begin{tabular}{lllllll}
    \toprule
     & cora & citeseer & cs & physics & computers & photo \\
    \midrule
    GCN* (sup) & 91.1$\pm$0.4 & 92.2$\pm$0.6 & 96.4$\pm$0.5 & 97.8$\pm$0.1 & 98.5$\pm$0.1 & 98.9$\pm$0.0 \\
    GRACE random* & 88.3$\pm$2 & 86.3$\pm$4.2 & 96.1$\pm$0.3 & OOM & 95.1$\pm$1.1 & 98.1$\pm$0.1 \\
    BGRL random* & 91.1$\pm$0.8 & 93.4$\pm$0.9 & 95.9$\pm$0.2 & 96.1$\pm$0.2 & 96.9$\pm$0.2 & 98$\pm$0.0 \\
    T-BGRL random* & 91$\pm$0.5 & 95.3$\pm$0.3 & 95.6$\pm$0.6 & 96.3$\pm$0.1 & 97.6$\pm$0.1 & 98.2$\pm$0.0 \\
    \midrule
    GCN (sup) & 92.75$\pm$0.76 & 90.78$\pm$1.04 X & 97.72$\pm$0.1 & \color{red}{98.88$\pm$0.04 $\bigstar$} & \color{blue}{98.83$\pm$0.06 $\bigstar$} & \color{red}{99.06$\pm$0.05 $\bigstar$} \\
    \midrule
    GRACE random & 94.22$\pm$0.61 & 95.9$\pm$0.62 & \color{violet}{\textbf{97.78$\pm$0.09}} & 97.96$\pm$0.08 & 98.67$\pm$0.04 & 98.91$\pm$0.06 \\
    GRACE deg & 92.99$\pm$0.7 & 94.63$\pm$0.6 & 97.62$\pm$0.11 & 97.71$\pm$0.06 & \color{violet}{\textbf{98.81$\pm$0.05 $\bigstar$}} & \color{violet}{\textbf{98.97$\pm$0.05}} \\
    GRACE evc & 94.46$\pm$0.75 & 94.0$\pm$0.8 & 97.45$\pm$0.1 & 97.4$\pm$0.09 & 98.66$\pm$0.04 & 98.88$\pm$0.05 \\
    GRACE pr & 93.15$\pm$0.67 & 95.23$\pm$0.39 & 97.71$\pm$0.08 & 97.8$\pm$0.08 & 98.76$\pm$0.05 & 96.73$\pm$0.26 X \\
    GRACE scom & \color{violet}{\textbf{94.69$\pm$0.5}} & 96.06$\pm$0.56 & 97.47$\pm$0.11 & \textbf{98.38$\pm$0.06} & 98.34$\pm$0.08 & 98.78$\pm$0.05 \\
    GRACE sbm & 91.52$\pm$1.21 & 95.31$\pm$0.68 & 97.53$\pm$0.13 & 98.01$\pm$0.09 & 98.15$\pm$0.05 X & 98.74$\pm$0.08 \\
    GRACE sbm2 & 94.26$\pm$0.41 & \color{red}{\textbf{96.36$\pm$0.47 $\bigstar$}} & 95.8$\pm$0.19 X & 98.32$\pm$0.04 & 98.5$\pm$0.04 & 98.47$\pm$0.09 \\
    \midrule
    CSGCL random & 92.27$\pm$0.85 & 95.0$\pm$0.69 & 96.76$\pm$0.15 & 98.07$\pm$0.05 & 98.74$\pm$0.04 & 98.78$\pm$0.08 \\
    CSGCL deg & 94.26$\pm$0.59 & 95.33$\pm$0.71 & 97.32$\pm$0.11 & \color{violet}{98.46$\pm$0.06} & 98.62$\pm$0.05 & 98.87$\pm$0.06 \\
    CSGCL evc & 93.79$\pm$0.93 & 93.5$\pm$0.72 & 97.64$\pm$0.1 & 98.28$\pm$0.06 & \color{red}{\textbf{98.88$\pm$0.04 $\bigstar$}} & 98.67$\pm$0.05 \\
    CSGCL pr & 94.2$\pm$0.9 & \textbf{96.24$\pm$0.52 $\bigstar$} & 97.43$\pm$0.08 & \color{blue}{\textbf{98.62$\pm$0.06 $\bigstar$}} & 98.52$\pm$0.05 & \textbf{98.95$\pm$0.05} \\
    CSGCL scom & 92.37$\pm$0.93 & 94.79$\pm$0.59 & \color{red}{\textbf{97.87$\pm$0.08 $\bigstar$}} & 98.02$\pm$0.07 & 97.66$\pm$0.08 X & 98.87$\pm$0.06 \\
    CSGCL sbm & \textbf{94.29$\pm$0.66} & 93.75$\pm$0.84 & 96.9$\pm$0.12 & 98.26$\pm$0.07 & 98.48$\pm$0.05 & 98.66$\pm$0.05 \\
    CSGCL sbm2 & 92.77$\pm$1.04 & 95.57$\pm$0.54 & 96.23$\pm$0.17 X & 96.94$\pm$0.11 X & 98.32$\pm$0.04 & 98.51$\pm$0.06 \\
    \midrule
    BGRL random & \color{blue}{94.97$\pm$0.6 $\bigstar$} & \textbf{95.37$\pm$0.62} & 97.7$\pm$0.07 & 98.3$\pm$0.06 & 98.6$\pm$0.06 & 98.46$\pm$0.06 \\
    BGRL deg & 94.55$\pm$0.59 & 91.08$\pm$1.57 X & 97.26$\pm$0.16 & 98.33$\pm$0.05 & 98.57$\pm$0.04 & \textbf{98.85$\pm$0.07} \\
    BGRL evc & \color{red}{\textbf{95.24$\pm$0.53 $\bigstar$}} & 92.97$\pm$0.87 & \color{blue}{\textbf{97.79$\pm$0.1}} & 97.8$\pm$0.08 & \textbf{98.68$\pm$0.04} & 98.82$\pm$0.07 \\
    BGRL pr & 92.11$\pm$0.98 & 94.7$\pm$0.71 & 96.82$\pm$0.12 & 97.85$\pm$0.07 & 98.65$\pm$0.06 & 98.75$\pm$0.05 \\
    BGRL scom & 92.37$\pm$1.15 & 91.85$\pm$1.03 & 97.52$\pm$0.13 & \textbf{98.35$\pm$0.05} & 98.58$\pm$0.05 & 98.5$\pm$0.07 \\
    BGRL sbm & 94.35$\pm$0.4 & 94.75$\pm$0.74 & 97.05$\pm$0.12 & 97.78$\pm$0.06 & 98.42$\pm$0.05 & 98.67$\pm$0.08 \\
    BGRL sbm2 & 93.28$\pm$0.78 & 93.16$\pm$0.62 & 96.48$\pm$0.09 & 97.8$\pm$0.1 & 98.38$\pm$0.04 & 98.38$\pm$0.08 \\
    \midrule
    L-GRACE random & 91.34$\pm$0.97 X & 95.7$\pm$0.64 & OOM & OOM & OOM & OOM \\
    L-GRACE deg & 93.95$\pm$0.9 & 94.03$\pm$0.75 & OOM & OOM & OOM & OOM \\
    L-GRACE evc & 90.72$\pm$1.01 X & 95.04$\pm$0.45 & OOM & OOM & OOM & OOM \\
    L-GRACE pr & 94.21$\pm$0.7 & \color{violet}{\textbf{96.26$\pm$0.59 $\bigstar$}} & OOM & OOM & OOM & OOM \\
    L-GRACE scom & \textbf{94.3$\pm$0.42} & 92.82$\pm$0.87 & OOM & OOM & OOM & OOM \\
    L-GRACE sbm & 92.94$\pm$0.79 & 92.51$\pm$0.63 & OOM & OOM & OOM & OOM \\
    L-GRACE sbm2 & 92.93$\pm$0.65 & 91.51$\pm$1.04 X & OOM & OOM & OOM & OOM \\
    \midrule
    L-BGRL random & \textbf{94.3$\pm$0.93} & \color{blue}{\textbf{96.3$\pm$0.48 $\bigstar$}} & 97.59$\pm$0.07 & \textbf{98.37$\pm$0.04} & 98.6$\pm$0.04 & 98.9$\pm$0.11 \\
    L-BGRL deg & 91.69$\pm$0.95 & 95.2$\pm$0.6 & 97.55$\pm$0.08 & 97.86$\pm$0.05 & 98.52$\pm$0.05 & 98.74$\pm$0.09 \\
    L-BGRL evc & 94.13$\pm$0.48 & 94.08$\pm$0.82 & \textbf{97.6$\pm$0.11} & 98.06$\pm$0.08 & 98.33$\pm$0.08 & \color{blue}{\textbf{99.06$\pm$0.06 $\bigstar$}} \\
    L-BGRL pr & 94.08$\pm$0.54 & 95.11$\pm$0.83 & 97.18$\pm$0.05 & 98.02$\pm$0.06 & \textbf{98.65$\pm$0.04} & 98.91$\pm$0.06 \\
    L-BGRL scom & 92.05$\pm$0.91 & 95.52$\pm$0.79 & 97.44$\pm$0.13 & 97.55$\pm$0.09 & 98.17$\pm$0.1 & 98.48$\pm$0.07 \\
    L-BGRL sbm & 93.89$\pm$0.56 & 95.22$\pm$0.55 & 97.19$\pm$0.18 & 97.29$\pm$0.05 X & 98.52$\pm$0.05 & 98.24$\pm$0.19 \\
    L-BGRL sbm2 & 93.96$\pm$0.4 & 95.76$\pm$0.58 & 97.21$\pm$0.09 & 97.53$\pm$0.1 & 98.47$\pm$0.05 & 97.21$\pm$0.42 X \\
    \bottomrule
    \end{tabular}
    }
    \label{tab:attr AUC}
\end{table*}
\end{document}